\documentclass[a4paper,fleqn]{cas-dc}
\usepackage[compress]{natbib}

\usepackage{pifont}
\usepackage{caption}
\usepackage{subfig}

\usepackage{times}
\usepackage{epsfig}
\usepackage{graphicx}
\usepackage{amsmath}
\usepackage{amssymb}
\usepackage{booktabs}
\usepackage{multirow}
\usepackage{pifont}
\usepackage{booktabs} 
\usepackage{multirow}
\usepackage{svg}
\usepackage[dvipsnames]{xcolor}  
\usepackage{amsmath}
\usepackage{nicefrac}
\usepackage[linesnumbered,ruled,vlined]{algorithm2e}
\usepackage{algpseudocode}
\usepackage[toc,page]{appendix} 
\usepackage{newunicodechar}
\newunicodechar{≥}{\ensuremath{\ge}}

\def\tsc#1{\csdef{#1}{\textsc{\lowercase{#1}}\xspace}}
\tsc{WGM}
\tsc{QE}
\tsc{EP}
\tsc{PMS}
\tsc{BEC}
\tsc{DE}

\begin{document}
\let\WriteBookmarks\relax
\def\floatpagepagefraction{1}
\def\textpagefraction{.001}

\shorttitle{Improving Prototypical Parts Abstraction for Case-Based Reasoning Explanations}

\shortauthors{D. Flores-Araiza et~al.}

\title [mode = title]{
Improving Prototypical Parts Abstraction for Case-Based Reasoning Explanations Designed for the Kidney Stone Type Recognition
}        
\author[addr1]{Daniel Flores-Araiza}
\author[addr1,addr2]{Francisco Lopez-Tiro}
\author[addr2, addr3]{Clément Larose}
\author[addr1]{Salvador Hinojosa}
\author[addr4]{Andres Mendez-Vazquez}
\author[addr1]{Miguel Gonzalez-Mendoza}
\author[addr1]{Gilberto Ochoa-Ruiz*}
\author[addr2]{Christian Daul*}

\cortext[cor]{Corresponding author(s): Gilberto Ochoa-Ruiz, Christian Daul \\ gilberto.ochoa@tec.mx, christian.daul@univ-lorraine.fr}

\address[addr1]{Tecnologico de Monterrey, Escuela de Ingenieria y Ciencias, Mexico}
\address[addr2]{Centre de Recherche en Automatique de Nancy (CRAN, UMR 7039), Universit\'e de Lorraine and CNRS, Vand{\oe}uvre-l\`es-Nancy, France}
\address[addr3]{CHRU de Nancy, Service d’urologie de Brabois, Vand{\oe}uvre-l\`es-Nancy, France}
\address[addr4]
{Centro de Investigación y de Estudios Avanzados, Computer Sciences Department, Guadalajara, Mexico}

\maketitle
\begin{abstract} 
 The in-vivo identification of the kidney stone types during an ureteroscopy would be a major medical advance  in urology, as it could reduce the time of the tedious renal calculi extraction process, while diminishing infection risks. Furthermore, such an automated procedure would make possible to prescribe anti-recurrence treatments immediately. Nowadays, only few experienced urologists are able to recognize the kidney stone types in the images of the videos displayed on a screen during the endoscopy. This visual recognition by urologists is also highly operator dependent.     
Thus, several deep learning (DL) models have recently been proposed to automatically recognize the kidney stone types using ureteroscopic images. However, these DL models are of black box nature and do not establish the relationship of the visual features they used to take the decision with the color, texture and morphological features visually analysed in biological laboratories to determine the type of extracted kidney stone fragments using the reference morphoconstitutional analysis (MCA) procedure. 
This contribution proposes a case-based reasoning DLmodel which uses prototypical parts (PPs) and generates local and global descriptors. The PPs encode for each class (i.e., kidney stone type) visual feature information (hue, saturation, intensity and textures) similar to that used by biologists during MCA. The PPs are optimally generated due a new loss function used during the model training. Moreover, the local and global descriptors of PPs allow to explain the decisions (``what'' information, ``where in the images'') in an understandable way for biologists and urologists. 
The proposed DL model has been tested on a database including images of the six most widespread kidney stone types in industrialized countries. The overall average classification accuracy was $90.37\pm 0.6 \%$. When comparing this results with that of the eight other DL models of the kidney stone state-of-the-art, it can be seen that the valuable gain in explanability was not reached at the expense of accuracy which was even slightly increased with respect to that ($88.2\pm 2.1 \%$) of the best method of the literature.
These promising and interpretable results also encourage urologists to put their trust in AI-based solutions.

\end{abstract}

\section{Introduction \label{Introduction}}
\subsection{Medical context \label{Intro-MCA}}

Urolithiasis (i.e., renal calculus formation) is a worldwide issue (\cite{quhal2021guideline}) entailing large expenditures on health systems (\cite{roberson2020economic}).  As reported in (\cite{kasidas2004renal}), urolithiasis affects at least 10\% of the population in industrialized countries and the risk of recurrence reaches up to 40\% in North America.

Kidney stones are aggregations of crystals that form in the urine. When their diameter becomes large (a few millimeters), kidney stones can remain blocked in the urinary tract (e.g., in a kidney calyx or a ureter) and cause severe pain. Kidney stones are classified into seven main types and twenty three sub-types (each type includes a given number of sub-types) according to their crystalline structure and biochemical composition. The formation causes depend on numerous risk factors such as patient genetics, age, weight, and sex, as well as the environment (warm or cold climate), lifestyle, comorbidity, or iatrogenic infections. A detailed description of the kidney stone types and sub-types, as well as the etiology (i.e., the causes of urolithiasis) can be found in (\cite{cloutier2015}).

Ureteroscopes (flexible endoscopes with a CCD matrix and optics on their distal tip) are used to display kidney stones on a screen. An optical fiber passing through the operative channel of the endoscope allows urologists to irradiate kidney stones using laser light pulses. The stones are then fragmented with an appropriately adjusted laser energy and pulse frequency. The fragments are ex\-trac\-ted and analyzed in a biology laboratory to determine their type and sub-type using a reference procedure referred to as morpho-constitutional analysis (MCA) \cite{daudon2016}.   

\begin{figure*}[t]
\centering
\includegraphics[width=0.8\linewidth]{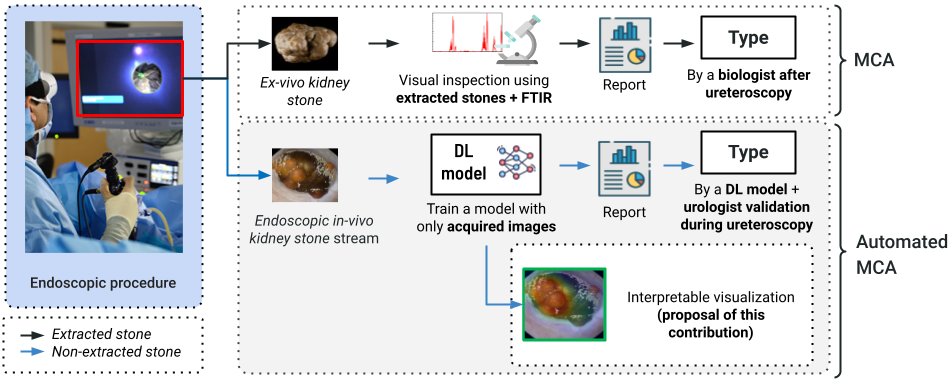}
    \caption{Overview of two procedures for determining the type of kidney stones. MCA is an ex-vivo procedure since it requires the extraction of kidney stone fragments from the urinary tract. In Morpho-Constitutional Analysis (MCA), fragments are analyzed by a biologist who determines the type of the kidney stones by a visual inspection followed by a biochemical (FTIR) analysis. On the other hand, automatic MCA (aMCA) uses machine learning-based methods to identify the type of kidney stones using endoscopic images acquired during the ureteroscopy (i.e., in in-vivo). A deep learning (DL) model is trained and exploited to perform real-time inference to assist the urologist in kidney stone recognition. In this contribution, the aMCA  is based on explainable Artificial Intelligence (XAI) models that allow to understand the decisions taken by the DL models.}
\label{fig:guidelines}
\end{figure*}

MCA is a two-step procedure (see on the top right side on Fig.~\ref{fig:guidelines}). First, the aspect of the kidney stone fragment surfaces and sections are visually observed with a mi\-cros\-cope. In this step, biologists describe the morphology of the crystal agglomeration using standardized key-terms relating to the colors, textures, and structure topology visible on the fragment surfaces and sections. This morphology analysis allows to recognize some types (i.e., crystal types as for instance whewellite, weddellite, and uric acid corres\-pon\-ding to types I, II and III, respectively) and some sub-types (as struvite, brushite or cystine denoted by IV.c, IV.d and V.a, respectively). In the second step, the fragments are powdered, and the spectra obtained by a Fourier Transform Infrared Spectroscopy (FTIR) gives the biochemical composition of the kidney stones.
This constitutional information is required to identify the remaining types and sub-types that cannot be distinguished using solely the morphological analysis. MCA is a reliable solution for recognizing kidney stone types and their sub-types (\cite{corrales2021classification}).   

However, even if the MCA is currently the reference solution for identifying kidney stones, this method also has its drawbacks. On the one hand, the MCA has to be performed ex-vivo and therefore requires the extraction of the kidney stone fragments from the urinary tract, which is a long and tedious task. This extraction process usually takes at least half an hour and involves the risk of infection. On the other hand, the biology laboratories in the majority of hospitals are not only in charge of the identification of kidney stones, but they are also responsible for the analysis of other tissues in the frame of various pathologies. For this reason, the kidney stone identification results are often only available after some weeks (\cite{turk2016eau}), whereas for some renal calculus types (e.g., with a very short recidivism time of some days) an immediate diagnosis and treatment is strongly recommended. 

Therefore, automated methods for in-vivo identi\-fi\-ca\-tion (i.e., performed inside the urinary tract) using the images acquired with an endoscope and displayed during the ureteroscopy would be an important step towards a significant improvement, both in terms of the endoscopic procedure duration and the anti-recurrence treatment definition time. On the one hand, the ureteroscopy duration can be significantly reduced since the renal calculus fragments can be pulverized (by adjusting appropriately the laser energy and pulse frequency) instead of extracting them. On the other hand, an automated image-based recognition method would favor a ``real-time'' diagnosis (i.e., during the ureteroscopy) for a rapid anti-recidivism treatment.

It must be noted that the results of the kidney stone type identification performed in biology laboratories (most often with MCA) were systematically used as ground truth to assess the performance of the classification algorithms described in the coming state-of-the-art section.

\subsection{Automated kidney stone identification}

Given its diagnostic importance, several authors have dealt with automated kidney stone identification. Some initial attempts were first made using shallow-based (i.e., classical) machine learning methods (\cite{SERRAT201741, martinez2020towards}), which led to promising results. Nonetheless, deeplearning-based (DL-based) approaches have rapidly become the favored approach to classify renal calculi (see the right bottom part of Fig.~\ref{fig:guidelines}).  

The first DL solution in the literature (\cite{black2020deep}) was based on a ResNet-101 architecture designed to recognize five kidney stone sub-types. The network weights were pre-trained with ImageNet and fine-tuned with kidney stone images acquired in ex-vivo under controlled acquisition conditions (i.e., in an enclosed environment with a diffuse and homogeneous scene illumination, and with well-chosen camera positions). Since only a limited number of images were available in this work, the authors augmented the data by manually extracting patches from the images, the size of the patches being chosen to capture enough texture and color information to allow for class separation. Encouraging results were reached by this precursor work since the five sub-types of kidney stones were classified with an acceptable overall performance (the recall, specificity, and precision values were 94.40\%, 96.40\%, and 82.11\%, respectively). Although this work highlighted the potential of DL approaches to identify kidney stones, in-vivo data (images acquired in the urinary tract and use of an endoscope instead of a conventional CCD camera) are by far more challenging than ex-vivo data captured from controlled viewpoints and without strong illumination changes and specular reflections.       

\begin{figure*}[bth] 
    \centering
    \includegraphics[width=0.70\linewidth]{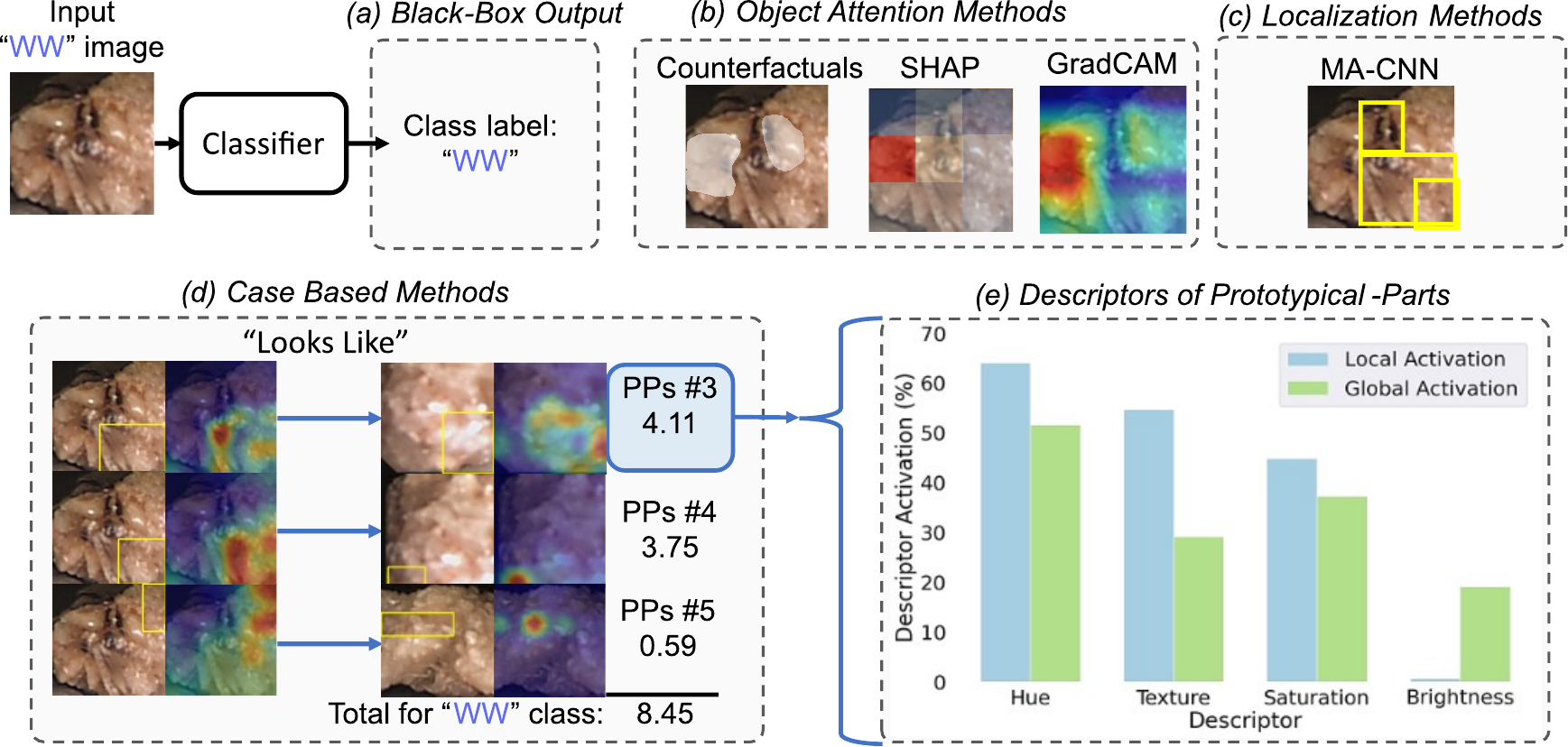}
    \caption{\small{Illustration of various interpretability levels in the frame of kidney stone type recognition (type ``WW'' stands for whewellite).  (a) Traditional non-interpretable  DL models produce a class label without any explanation.
    (b) Three most common object-level attention maps: i) counterfactual explanation (\cite{Guidotti2022}) which indicates for each image region the smallest change in feature values that can modify the class label, ii) SHapley Additive exPlanations (SHAP, (\cite{Lundberg2017})) which is based on game theory and iii) GradCAM (\cite{GradCAM}). (c) Decision region delineated with a localization method  (\cite{Fine_Grained_Zheng_2017_ICCV}), (d) Case-based methods quantify the similarity of ``decision regions'' and prototypical parts (PPs, (\cite{ProtoPnet})). 
    (e) PP-descriptors give an indication on the important local and global image features (here hue, saturation, and intensity values in the HSI color space (\cite{daul2000}) and LBP texture histograms (\cite{SERRAT201741}).
    }
    \label{fig_XAI_methods}}
\end{figure*}

In (\cite{estrade2022towards}), the authors considered three sub-types with different biochemical compositions, namely sub-types Ia (calcium oxalate monohydrate), IIb (calcium oxalate dihydrate) and IIIb (uric acid), the aim of this contribution being to identify kidney stones which can belong to one of five classes (three classes of pure kidney stones of sub-types Ia, IIb and IIIb and two classes of mixed stone compositions of sub-types Ia+IIb and Ia+IIIb). The images were gathered in two datasets, that of the kidney stone fragment surface images and that of the fragment section images. Data augmentation was also performed by applying geometrical transformations (i.e., a combination of translations, scaling and rotations) on the images. Two ResNet-152-V2 architectures were trained to classify the kidney stones either only with surface data or only with section data. While for the five classes taken individually, the specificity is constantly high (at least 90\%), and the recall values are very different (from 50\% up to 98\%), the overall percentage (percentage over the five classes) of correct renal calculus identification is rather satisfactory, both for surface (83\%) and for section (81\%) data. 

However, the separate use of surface and section data is an obstacle for improving the efficiency of kidney stone recognition. Indeed, a contribution (\cite{lopez2024vivo}), which compares the efficiency of the main kidney stone identification approaches based on in-vivo images, has shown that, when a model is trained by simultaneously using surface and section data, the performances can be improved both by well-tuned shallow-based machine learning and DL approaches. One of the most significant improvements with data fusion was reported in (\cite{Lopez-Tiro_2023_ICCV}). The latter reports an accuracy increase of 11\% over five classes when attention and multi-view feature fusion strategies are used instead of single views.

\subsection{Scientific motivation and paper structure}
A decision requires a justification in all medical applications. In urology, an anti-recurrence treatment of lithiasis is supported by the MCA report, which explains the decision made by humans. 
Some attempts have been made for shallow-based machine learning to understand the meaningful decision features (e.g., discriminant features in appropriate color spaces (\cite{martinez2020towards}) or efficient texture representations (\cite{SERRAT201741})) for identifying kidney stones. Shallow-based machine learning has the advantage that the classification exploits physically interpretable features. However, this interpretability comes at the cost of a lower accuracy.

On the other hand, the limitations of DL architectures lie on their ``black box'' nature (\cite{RISE_Petsiuk2018}) since the training of millions (or even billions) of parameters allows for high performance in terms of accuracy at the expense of erroneous decisions that cannot be associated to the incorrect set of values used by the neural network.  

The field of explainable AI (XAI) seeks to provide AI systems with descriptions of their rationale and decision-making process. XAI methods have helped for instance to characterize the reasons behind a model performance and to assess the appropriateness of the model and the training data,  thus enabling to build trust in DL models. 
Consequently, XAI enables a responsible approach to AI development and even facilitates debugging and improvement of AI models (\cite{ProtoDebug_bontempelli2023, AlvarezMelis2018TowardsRI}).
Figure \ref{fig_XAI_methods} gives an overview of how XAI could be used in the context of kidney stone classification.  Figure \ref{fig_XAI_methods}.(b) shows three saliency map types indicating where in its input image a model mainly relies on to take a decision. Bounding boxes of the salient region of interest (see Fig.~\ref{fig_XAI_methods}.(c)) can be another way to indicate important decision regions in images. 
Even though these classical XAI methods provide a foundation for interpretability in DL models, they are still insufficient for complex recognition tasks using Computer Aided Diagnosis (CAD) systems. 

Recent XAI methods give more precise explanations on the decisions taken by DL models, typically after the labels were assigned to the input data. 
A holistic explanation provides a description of \textit{what}, \textit{where}, and \textit{why} visual features are relevant. Quantitative evaluations of visual features as those given in Fig.~\ref{fig_XAI_methods}.(e) are an indication \textit{why} a given feature is helpful, while ground truth-based saliency maps can precisely highlight \textit{what} feature is relevant and \textit{where} it is located within the image (see  Fig.~\ref{fig_XAI_methods}.(d)). 

This contribution aims to improve the interpretability and accuracy of self-explainable models for kidney stone identification using a case-based reasoning approach based on Prototypical Parts (PPs). 
The model should not only ac\-cu\-ra\-te\-ly identify kidney stone types but also provide transparent and understandable explanations for its decisions, which is critical in this medical application where trust and clarity are essential. By using PPs, the model may create case-based reasoning explanations that are consistent with how biologists recognize kidney stones, thus allowing for better decision-making and clinical acceptance.

The rest of this paper is organized as follows.
Section \ref{XAI_SOTA} presents the current trends in XAI and discusses their limits for image classification. This section focuses on self-explainable methods and justifies the potential of PP-based models for kidney stone identification. 
Section \ref{Proposed_Framework} details the proposed DL solution, which is based on a modification of the loss function of ProtoPNet (a PP-based explainability network (\cite{ProtoPnet}). 
Section \ref{Experimental_Setup} presents the experimental set-up, which includes the used kidney stone datasets, as well as the different model configurations and methods used to evaluate their efficiency.
Section \ref{Results} gives both a quantitative and qualitative result analysis of kidney stone identification. It discusses also how the proposed model can be efficiently used.
Finally, sections \ref{Discussion} and \ref{Conclusion} respectively recall the main paper contributions and provide a conclusion.

\section{Recent XAI advances in image processing \label{XAI_SOTA} }

Attempts to explain the decisions of DL models can be divided into post-hoc and self-explainable methods (\cite{XAI_fieldGuide}). 
%
%
In the former category, the behavior of a model is systematically observed after its training, for instance, by analyzing the model responses concerning input modifications (see Fig.~\ref{fig_XAI_methods}.(b)). This first category also includes approaches that generate saliency maps using the inner states and weights of the model (\cite{RISE_Petsiuk2018, Shap2017}). Other methods provide counterfactual examples highlighting minimal input alterations needed to change the model's output. The  ``added value'' of this latter approach is that it does not only indicate \textit{what} feature tends to change the class label but also \textit{how much} the feature value must vary to modify the output (\cite{jeanneret2023adversarial}). However,  although they are easy to implement, post-hoc explanations can be biased and unreliable (\cite{Adebayo2018SanityCF}).

In contrast, “self-explanatory” models are designed to make their decision-making transparent (\cite{Brendel2019ApproximatingCW, AlvarezMelis2018TowardsRI}). 
These methods provide insights into the internal behavior of models through concepts easily and utilized by domain experts, such as concept activation vectors (\cite{Kim2017TCAV, Chen2020ConceptWF}) or model attention and activation spaces for explanations with adversarial auto-encoders (\cite{guyomard2022vcnet}). 
Recently, an increasing number of self-explainable approaches have been built on ProtoPNet (\cite{ProtoPnet}). This network configures the activation space by learning a hidden layer of prototypical parts (PP) given the activation patterns learned by the convolutional layers of the model.  
When faced with challenging recognition tasks, human experts often try to define decision rules by searching to localize in sub-image regions specific prototypical aspects characterizing the classes.

However, PP-based methods can still suffer from ambiguity between the learned parts since it can be challenging to define what constitutes a ``part'' for some classes.
Furthermore, what a DL model considers as a PP might differ from human perception. 
For instance, it has been shown in (\cite{Flores_Araiza_2023_CVPR}) that ProtoPNet can identify many classes using visually analogous features, making it difficult for clinicians to build trust in the network's classifications.

\subsection{Self-explainable methods \label{Self_explain_methods}}

ProtoPNet resulted from pioneer work in the field of PPnetworks. The concept of interpretable prototypes allowed to improve the understanding of the decisions taken by image classification models. 
These prototypes, learned from the model's latent space, are refined during model training to closely reflect the training data. Knowing such prototypes enables a direct and understandable explanation of the decisions taken by deep neural networks (DNNs) while maintaining their performance. 
ProtoPNet has inspired the design of numerous self-explainable models. 
For instance, TesNet (\cite{tesNet_Wang_2021_ICCV}) constructs the latent space on a Grassman manifold, without considering the number of PPs required for each class. 
Conversely, ProtoPool (\cite{ProtoPool_Rymarczyk_2021}) and ProtoTree (\cite{PPs_Trees_Nauta_2021_CVPR})  were both conceived to reduce the number of prototypes needed for inference: ProtoPool employs a differentiable assignment strategy to semantically merge similar prototypes, whereas ProtoTree organizes prototypes into a binary decision tree to combine global interpretability with local explanation capabilities.
The extension of part-prototype networks into areas such as deep reinforcement learning (PW-Net in (\cite{PWNet_kenny2023towards}), and model debugging (ProtoPDebug in (\cite{ProtoDebug_bontempelli2023}), highlights the adaptability of the method and the broad interest on this approach.

\subsection{Limitations of current PP-based XAI methods\label{Self_explain_limitations}} 
Case-based reasoning architectures like ProtoPNet tend to produce very similar PPs, leading to a collapse with just a few training images, especially in datasets with a limited number of samples (\cite{Flores_Araiza_2023_CVPR}). This behavior, similar to the mode collapse observed in GANs (\cite{bau2019seeing}), is particularly an issue in medical diagnosis, which requires fine-grained differentiation between classes. Further, this issue may impair the model's ability to recognize subtle distinctions, risking overfitting and poor generalization.
A high similarity among PPs is another issue since it reduces the diversity of informative features, making the interpretability less meaningful for the specialist (urologists or biologist in the context of our work) utilizing them. Moreover, a semantic gap exists between similarities in the latent and input spaces, particularly under strong photometric perturbations (as occurring in endoscopy), where PPs do not align with human prioritization of visual features (\cite{hoffmann2021looks}). Additionally, most cases of non-human aligned PPs have been found in erroneous classification cases (\cite{PP_Descriptors}). 

This work explores appropriate modifications of the loss functions in a ProtoPNet implementation to counteract the issues of prototype homogeneity and semantic ambiguity. For instance, (\cite{PP_Descriptors}) makes an analysis of the prototypes under various realistic photometric perturbations. These perturbations, naturally occurring according to the image domain task, serve to clarify the meaning of PPs by quantifying the influence of visual characteristics relating to textures or hue and saturation values in the HSI color space (\cite{daul2000}).
Our approach adopts the term ``descriptor'' for these additional prototype characterizations used to assess the significance of visual features in relation to the identified prototypes. To sum up, this contribution uses descriptors to quantify the relevance of specific visual attributes to the learned prototypes of the model. This quantification allows to refine the interpretability of AI models by classifying images based on prototypical components.
In particular, knowing the contribution of the descriptors to the class attribution enables highlighting the trade-off between interpretability and performance in complex medical applications in which human-aligned interpretability is required. 

\subsection{Contributions and overview of this work}\label{Contributions_of_this_work}
\begin{figure*}[t]
\centering
    \includegraphics[width=1.0\linewidth]{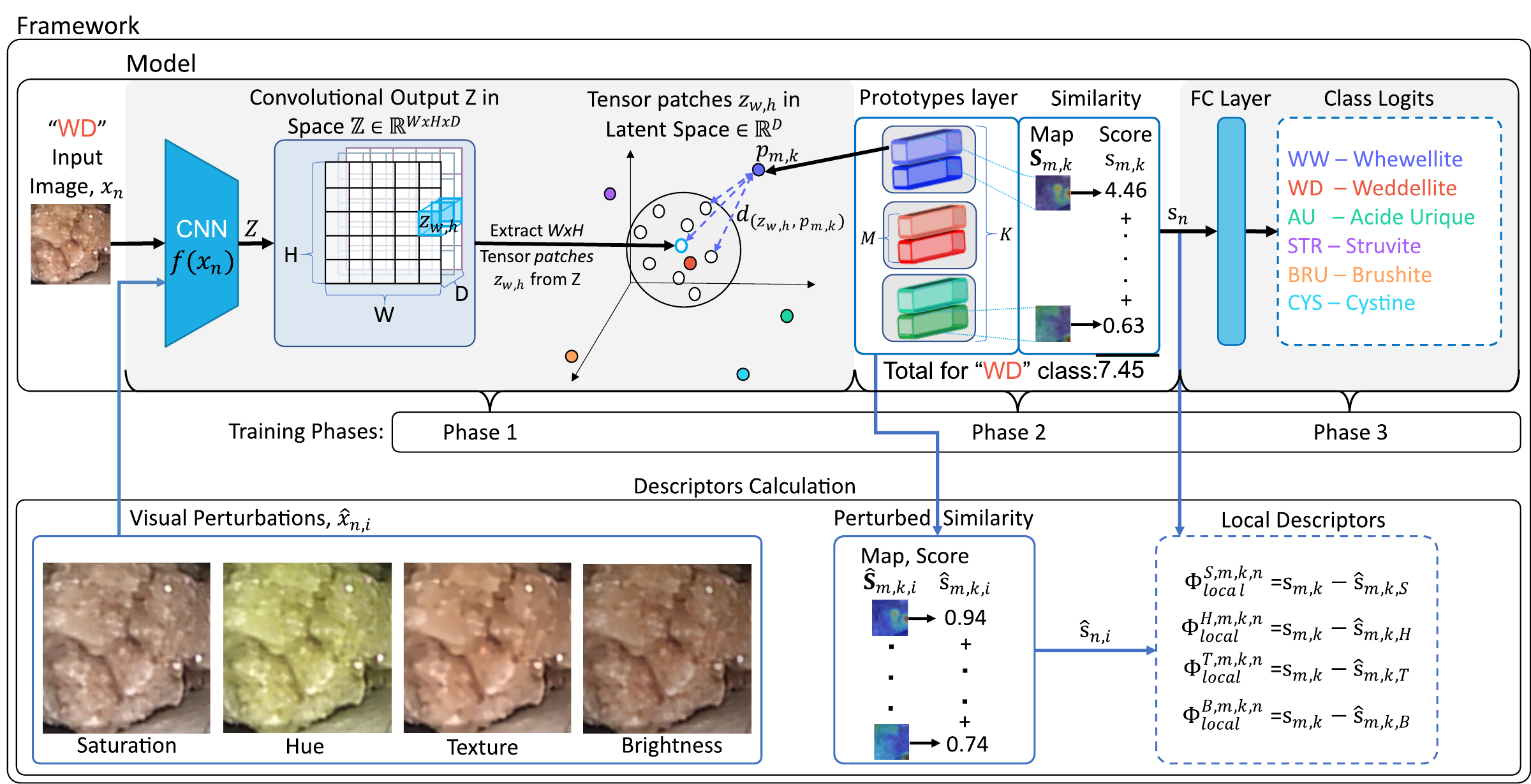} 
    \caption{\small{\textbf{Proposed DL model.} Input image $x_n$, an acquisition of a weddellite (WD) kidney stone in this example, is processed by CNN $f$ that translates the content of $W \times H$ image patches to a latent feature representation whose discrete volume dimension is given by $\mathbb{R}^{W\times H\times D}$, $W$ and $H$ being respectively the number of the adjacent patches along the columns and lines of $x_n$ (the use of image patches is justified in Section \ref{Kidney_stone_dataset}). As sketched in this figure, the tensors $z_{w,h}$ of each image patch correspond to a point (white discs) in an embedding space of 128 dimensions ($D$=128).
    In this space, the $L2$-distances $d_{z_{w,h},\,p_{m,k}}$ between tensors $z_{w,h}$  and learned PPs (tensors $p_{m,k}$, colored discs) are all assessed. These $L2$-distances are used to compute similarity maps $\mathbf{S}_{m,k}$  (see the ``similarity block'' of this figure and the map examples given in Fig.~\ref{fig_XAI_methods}.(d)) which allow to quantify the resemblance of a PP and an image patch. The greatest value of a similarity map acts as similarity score $s_{m,k}$, which indicates to which input image patch the PP is the closest. Finally, the similarity scores $s_{m,k}$ of all PPs are processed by a fully connected (FC) layer to get the logits. A softmax is applied to the logits to determine the class label.} \label{fig_general_architecture}}
\end{figure*}

This work aims to reduce the limitations of DL models and XAI-methods by providing comprehensive visual explanations in the context of kidney stone type identification. Traditional XAI methods, often relying on visual explanations based only on heatmaps, tend to oversimplify the complex process of classifying kidney stones into specific types.
This paper introduces a novel approach that improves the interpretability and effectiveness of DL models used as computer-aided diagnostic tools. As illustrated by the overview in Fig.~\ref{fig_general_architecture}, the proposed DL architecture
extracts semantic features 
from an input image using a Convolutional Neural Network (CNN). These features are then compared to those extracted from learned PPs. The resemblance of the learned PPs and the image parts to be recognized is quantified by semantic feature similarity scores, the class labels being obtained by a weighted combination of similarity scores. This method ensures that the explanations are faithful to the model's inner behavior by using the same PPs for both the model output and its explanations. The proposed model limits the number of PPs to facilitate user understanding while achieving competitive performance against its non-interpretable counterpart. 
%
The generated explanations are based on descriptors similar to that standardly used during a MCA, i.e., the explanation tries to mimic the rules followed by biologists when they visually identify kidney stone types with the microscope.  This way to proceed is a first step towards clinical applicability and urologist acceptability (\cite{Flores_Araiza_2023_CVPR}).

However, recognizing kidney stone types in images requires high-level expertise. Indeed, biologists undergo training in centers dedicated to the recognition of kidney stones and must then gain experience over several months, or even years, to carry out the MCA described in Section \ref{Intro-MCA}. Only a few urologists are able to identify the kidney stones on a screen during a ureteroscopy. The explanations, even based on PPs covering small areas of the kidney stone images, rely on features that are difficult to analyze for non-experts. This contribution addresses this issue by quantifying and understanding the sensitivity of PPs to various perturbations (kidney stone aspect changes due to the endoscope's viewpoint, changing illumination, etc.).  

This approach produces easily understandable predictions for specialists, making the decisions of an automated kidney stone classification clear. It reproduces the morphological analysis part of MCA described in Section \ref{Intro-MCA}, building trust in the AI system, and allows for the adjustment of the model's output when specialists express this need.

The proposed DL architecture employs a case-based reasoning approach based on ProtoPNet, which is augmented by a Deep Metric Learning (DML) focused loss function to refine the embedding space of extracted features. This refinement enhances the discrimination power of PPs, thereby improving the overall accuracy and interoperability of the model.
A DML-focused loss function aids in optimizing the distances between embeddings by better guiding the definition of the decision space, thus enhancing the models' ability to measure the similarity to prototypical cases for each class. This enhancement enables the classification of the input images with an additional visual characterization of the reasons learned by the model to detect a similarity between the trained PPs and the visual features of the input image.  
The contribution also explores \textbf{i)} different CNN backbones, \textbf{ii)}  the required number of PPs per class, \textbf{iii)} the relevance of data augmentation in training, and \textbf{iv)} the impact on the results of various loss functions. Notably, the model training does not require any part annotations, relying solely on class labels.
The proposed approach, with its inherently interpretable reasoning process, contrasts directly with previous works that relied on post-hoc explanation techniques to explain a trained black-box model on particular classifications (\cite{lopez2024vivo, estrade2022towards}) or with global explanations (\cite{elbeze2022evaluation}).

\section{Proposed DL architecture\label{Proposed_Framework}}

This section starts with an overview of the proposed ProtoPNet-based solution's modus operandi. Then, it describes the model's training and highlights its limitations. Finally, it shows how these limitations can be overcome using an appropriate loss function to avoid the PP collapse of a ProtoPNet-based architecture.    

As sketched in Fig.~\ref{fig_general_architecture}, the proposed DL model is first trained to produce i) a set of useful weights in the feature extraction layers, ii) a set of PPs in the prototype layer, and iii) the weights of a fully connected layer which translates the similarity measured between the PPs and the visual features of an input image into a class label. Once trained, the model can be used for inference purposes.

\subsection{Components of the inference stage\label{Components_of_inference}}
The goal of the proposed DL architecture is to produce a classification based on a set of clear and comprehensible descriptors providing an explanation that is interpretable by biologists performing MCA and urologists making ureteroscopies. Explanations are based on learned PPs to reach this goal. After the training, the DL model is used for inference. This sub-section details the inference stage sketched in  Fig.~\ref{fig_general_architecture}. \\

\textbf{Prototypical Image Encoding: } 
The first stage of the inference pipeline deals with the encoding of the images into a set of feature activations. This is achieved through the use of a pre-trained CNN-backbone extracting semantic features from input image $x_n$.  With an appro\-pria\-te training,  this first feature extraction step should lead to diversity and representativity in terms of extracted features.
Three CNN-backbones $w_{base}$ were tested  (namely \textit{VGG16}, \textit{ResNet50}, and \textit{DenseNet201}) to evaluate their impact on the performance of the proposed approach. Two layers of $1\times1$ convolutions $w_{add}$ follow the extraction backbone $w_{base}$ to adjust the depth of the feature activation maps to a 128-channel depth (see Fig.~\ref{fig_general_architecture}). A CNN-backbone, together with the two layers of $1\times1$ convolutions, form feature extractor $f$. The latter is applied to input image $x_n$ so that $Z= f(x_n)$ generates the convolutional output $Z$ of $W\times H$ latent feature tensors in space $\mathbb{Z} \in \mathbb{R}^{W\times H \times D}$. The coordinates in the three-dimensional discrete latent feature space $\mathbb{Z}$ are $w$$\,\in$ [1, $W$=7], {\em h}$\,\in$ [1, $H$=7] and {\em d}$\,\in$ [1, $D$=128], where $w$ and $h$ define column and line numbers in a regular square grid of adjacent convolutional patches extracted from image $x_n$ (see Fig.~\ref{fig_general_architecture}). Thus, discrete space $\mathbb{Z}$ encodes $L = W\times H$ latent feature tensors $z_{w,h}$ of dimension $1\times1\times128$ and associated each with an image patch located on column $w$ and line $h$ of the patch grid. 
  
The learned prototypical-part (PPs) tensors $p_{m,k}$ are also of $1 \times1\times 128$ dimensions to enable their comparison with the latent feature tensors $z_{w,h}$.
As sketched in Fig.~\ref{fig_general_architecture}, in the proposed model, the PPs form a single layer referred to as the ``prototype layer''. 
This layer is based on $P = M\times K$ PPs since a constant number of $M$ prototypes are learned for each of the $K$ classes. The prototypes $p_{m,k}$ are indexed by $m$ and $k$, with $m$$\,\in$ [1, $M$] and $k$$\,\in$ [1, $K$].
The learned PPs are expected to be representative of the prototypical activation patterns of the class $k$ to which they belong. 
Thus, squared $L{2}$ distances  $d_{z_{w,h}, p_{m,k}}$ are determined in the latent space \(\left(\textrm{in} \, \mathbb{R}^{D}\right)\) for all combinations of the $p_{m,k}$ and $z_{w,h}$ tensors. These squared distances are  obtained with Eq.~(\ref{eq_distance})
\begin{equation}\label{eq_distance}
d_{z_{w,h},\,p_{m,k}} = \lVert  z_{w,h} - p_{m,k} \rVert_{2}^{2}
\end{equation}
and are used in Eq.~(\ref{eq_similarity_score}) to convert them into scores  $s_{w,h,m,k}$ quantifying the similarity of the PPs tensors $p_{m,k}$ and the $z_{h,w}$ patch tensors. 
\begin{equation}\label{eq_similarity_score}
s_{h,w,m,k} = \ln \left( \frac{d_{z_{h,w}, p_{m,k}}+1}{d_{z_{h,w}, p_{m,k}}+\epsilon}\right)
\end{equation}
\noindent In Eq.~(\ref{eq_similarity_score}), $\epsilon$ is a small value to avoid division by zero.

\begin{figure*}[t]
\centering
    \includegraphics[width=1.0\linewidth]{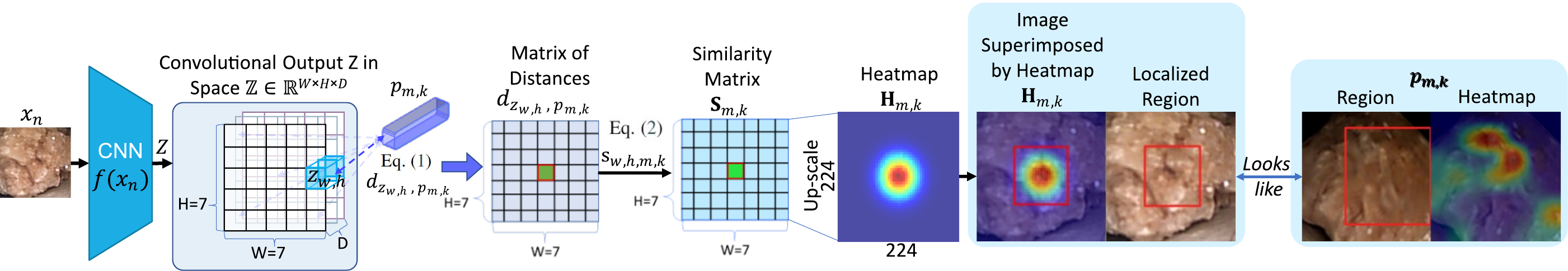} 
    \caption{\small{\textbf{Similarity visualization: } Superimposition of heatmap $\mathbf{H}_{m,k}$ of the most similar PP $p_{m,k}$ on the corresponding region in image $x_n$} \label{fig_PP_similarity_visualization}}
\end{figure*}

As noticeable in the ``map'' column of the ``similarity'' block in Fig.~\ref{fig_general_architecture}, map $\mathbf{S}_{m,k}$ is a matrix of similarity scores $s_{w,h,m,k}$ which encodes the similarity between a given $p_{m,k}$ tensor and all the latent feature tensors $z_{w,h}$ extracted from the $W\times H = 49$ patches of input image $x_n$ (i.e., $\mathbf{S}_{m,k}$ are matrices with dimension $7\times7$). Also, as maps $\mathbf{S}_{m,k}$ preserve the spatial arrangement of input image $x_n$, they can be upscaled (i.e., using bilinear interpolation) to produce heat maps $\mathbf{H}_{m,k}$, shown in Fig.~ \ref{fig_PP_similarity_visualization}. 
A global max-pooling operation is applied to each similarity map $\mathbf{S}_{m,k}$ to obtain the largest similarity score $s_{m,k}$ between a PP $p_{m,k}$ and all the latent feature patches $z_{w,h}$ extracted from input image $x_n$. This highest similarity score $s_{m,k}$ quantifies the best resemblance of a prototypical-part to a particular area (patch)  in $x_n$. 

Finally, the $P=M\times K$ highest similarity scores $s_{m,k}$ of all PPs-tensors $p_{m,k}$ are processed with a fully-connected (${FC}$) layer and a softmax function that provides $K$ class label probabilities $a_k$ as output.
Algorithm \ref{Algorithm_Framework_Classification_and_Explanations} gives an overview of the described PPs-based classification. \\ 
 
\begin{algorithm}[tb]
\DontPrintSemicolon
\KwIn{Set $\mathbf{X}$ of $N$ input images $x_n$, $K$ classes $k$, feature extractor $f$ (consisting of backbone $w_{base}$ and the two $1\times 1$ convolution layers $w_{add}$), $P=M\times K$ prototypical parts (i.e., $M$ PPs $p_{m,k}$ per class $k$), fully connected layer $\mathbf{W}_h$ (classifier) with bias $b_h$, $I$ visual color and texture features $i$ to disturb.}
\KwOut{Per input image $x_n$: $K$ probabilities of class labels ${a}_k$, $P$ heatmaps $\mathbf{H}_{m,k}$ and $I$ descriptors per PP $p_{m,k}$}

\For{each input image $x_n \in \mathbf{X} $}{From $x_n$, extract convolutional output $Z=f(x_n)$ forming space $\mathbb{Z} \in \mathbb{R}^{H\times W \times D}$.

Split convolutional output $Z$ into $H \times W$ latent feature patches $z_{w,h} \in \mathbb{R}^{D}$

\For{each prototypical part $p_{m,k}$}{
// Determination of similarity map $\mathbf{S}_{m,k}$:\;
\For{each latent features $z_{w,h}$}{
Compute $d_{z_{w,h}, p_{m,k}}$ with Eq.~(\ref{eq_distance})\;

Determine score $s_{w,h,m,k}$ using Eq.~(\ref{eq_similarity_score})\;

Update the similarity matrix of $p_{m,k}$: \( \mathbf{S}_{m,k}(w,h) = s_{w,h,m,k}\)
}
$s_{m,k} = \text{max-pooling}(\mathbf{S}_{m,k})$\;
$\mathbf{s_n(m+k*M)} = s_{m,k}$, with vector $\mathbf{s_n}$ of $x_n \in \mathbb{R}^{P}$\;
}
\textbf{// Visualization of similarity $p_{m,k}$:}{ \;
Get heatmap $\mathbf{H}_{m,k}$ by upscaling \( \mathbf{S}_{m,k}\)\;
Superimpose heatmap $\mathbf{H}_{m,k}$ on image \(x_n\)\;
}
\textbf{// Suggested image classification:}\;

$a_k = \text{softmax}(\mathbf{W}_h \cdot \mathbf{s_n} + b_h)$\;

\textbf{// Calculation of the local descriptors :}\;

\For{\text{each visual perturbation (\textit{descriptor})} $i \in I$}{
    $\hat{x}_n = {Perturbation_i}({x}_n)$ \;
    
    $\hat{Z}=f(\hat{x}_n)$  // Similarly to line 2\;
    Split convolutional output  $\hat{Z}$ into $H \times W$ latent feature patches $\hat{z}_{h,w} \in \mathbb{R}^{D}$
    
    \For{each prototypical part $p_{m,k}$}{
    Determine $\hat{s}_{m,k,i}$ with lines 6 to 10 and by using $\hat{z}_{h,w}$, $d_{\hat{z}_{w,h},p_{m,k}}$ and $\hat{s}_{w,h,m,k}$ in Eqs.~(\ref{eq_distance}) and (\ref{eq_similarity_score}).\;
    $\Phi_{local}^{i,m,k,n} = s_{m,k} - \hat{s}_{m,k,i}$ (see Eq.~(\ref{Local_descriptors})) \;
    }
}
}
\caption{Principle of the image classification and generation of explanations with prototypical parts (PPs). Comments on algorithm parts start with symbols ``//''. 
}
\label{Algorithm_Framework_Classification_and_Explanations}
\end{algorithm}

\textbf{Visualization of PPs:}
A visualization of the PPs tensors $p_{m,k}$ learned using set $\mathbf{X}_{\textrm{train}}$ of \(\left|\mathbf{X}_{\textrm{train}}\right|\) training images $x_n$ \(\left(n \in \left[1, \left|\mathbf{X}_{\textrm{train}}\right|\right]\right)\) highlights the ability of the model to measure the similarity between image patches and prototype parts. This visualization provides detailed explanations that are crucial for the interpretability, trust, and reliability of the results provided by a DL model. Furthermore, it can support the process of model debugging, which can help DL experts improve the model capabilities.
In the practical case of ureteroscopy,  visualizing how the model associates certain parts of an image with learned PPs helps urologists and biologists to understand and learn the critical visual descriptors according to the kidney stone type. 
Furthermore, urologists' or biologists' ability to see which parts of an image contribute to kidney stone identification increases their trust in the AI application.

The squared $L2$ distance $d_{z_{w,h},\,p_{m,k}}=\lVert z_{w,h} - p_{m,k} \rVert_{2}^{2}$ quantifies the resemblance of prototype $p_{m,k}$ and tensor $z_{w,h}$ associated to the patch with coordinates $(w,h)$ in input image $x_n$. These $d_{z_{w,h},\,p_{m,k}}$-values are again used in Eq.~(\ref{eq_similarity_score}) to determine similarity scores $s_{w,h,m,k}$.  
$P=M\times K$ similarity matrices \( \mathbf{S}_{m,k} \) are obtained by successively determining the $d_{z_{w,h},\,p_{m,k}}$ and $s_{w,h,m,k}$ values  for all of the prototypes $p_{m,k}$ whose similarity is measured against the $W\times H = 49$ tensors $z_{h,w}$ of the convolutional patches of input image $x_n$.
Also, since these similarity matrices \( \mathbf{S}_{m,k} \) preserve the spatial arrangement of input image $x_n$, they are up-sampled using bilinear interpolation to get a heatmap $\mathbf{H}_{m,k}$ that will be superimposed on $x_n$ to visualize the similarity of $p_{m,k}$ and a region of $x_n$.
This superimposition is illustrated in Fig.~ \ref{fig_PP_similarity_visualization} and an example of  heatmaps for each PPs is given in the ``Map'' column of the ``similarity block'' in Fig.~ \ref{fig_general_architecture}. 
\\

\textbf{Descriptors calculation:}
The heatmap visualization method described in the previous paragraphs does not exhi\-bit a significant difference compared with the capabilities of existing post-hoc methods. However, this visualization can be improved using the descriptor calculation method described in (\cite{PP_Descriptors}). 
Descriptors help to highlight the significance of visual features for the model's ability to find PPs similar to the most relevant input image regions.
This approach is well aligned with the context of the MCA described in Section \ref{Intro-MCA}. Indeed, several contributions in ureteroscopy (\cite{corrales2021classification})  enforce the idea that significant visual features are present in several different areas of a kidney stone image. 
Therefore, the herein proposed descriptors allow for measuring the importance level of visual features for each PP.

The most informative image features in PPs can be detected by measuring changes in the similarity score when perturbing the input image with modifications so that value $s_{m,k}$ changes in $\hat{s}_{m,k,i}$. Index $i$ in the perturbed score  $\hat{s}_{m,k,i}$ denotes the type of modification (i.e., $i$ = S, H, T or B for saturation, hue, texture, or brightness, respectively, as sketched in the bottom of Fig.~\ref{fig_general_architecture}). 
$\Phi_{local}^{i,m,k,n}$ is the ``local'' importance score of feature $i$ when comparing the similarity of propototype tensor $p_{m,k}$  and test image number $n$ belonging to image set $\mathbf{X}_{\textrm{test}}$. The value of  $\Phi_{local}^{i,m,k,n}$ is given by the difference of two similarity scores measured with and without the perturbation:
\begin{equation}\label{Local_descriptors}
    \Phi_{local}^{i,m,k,n} = s_{m,k} - \hat{s}_{m,k,i}
\end{equation}
Through Eq.~(\ref{Local_descriptors}), the local descriptors provide a quantifiable estimation of how a photometric perturbation in an input image affects the resemblance of the latter and of the PPs. The descriptor leading to the highest local importance score $\Phi_{local}^{i,m,k,n}$ reveals the key visual feature that drives the similarity of the specific PP tensor $p_{m,k}$ being analyzed. 
Moreover, it is possible to determine the global significance of each visual perturbation for each PP descriptor. 
These global significance scores are referred to as ``global descriptors''. The estimation of global descriptors are detailed at the end of the following Section \ref{Training_procedure}.

\subsection{Training procedure\label{Training_procedure}}
As detailed in Algorithm \ref{Training_algorithm} and sketched in Fig.~\ref{fig_general_architecture}, the training procedure begins with an initialization followed by three sequentially chained loops (corresponding to three training phases), which are iterated $N_{\textrm tc}$ times (parameter $N_{\textrm tc}$ fixes the training cycle number). 

In \textit{Phase 1} of Algorithm \ref{Training_algorithm}, the convolutional layers of CNN-backbone $f$ are updated for $N_{\textrm f}\,=\,10$ epochs.
During this phase, the weights of the feature extractor are updated, with learning rate $\eta$, to capture the semantic features of the input images. These semantic features are encoded in $H\times W$ latent feature tensors $z_{w,h}$.

In \textit{Phase 2} of Algorithm
This phase ensures that PPs $p_{m,k}$ have a direct visual equivalence from a patch area in a training image of their respective class and enables their use for case-based reasoning explanations.

In \textit{Phase 3} of Algorithm \ref{Training_algorithm},  the weights of the FC-layer (i.e., the last layer of the model) are updated in $N_{\textrm h} = 20$ epochs. 
This phase aims to adjust the final classification layer so that the similarity scores accurately lead to the correct class labels.
The first and third training phases use a same loss function $\mathcal{L}$.

%
%
Convergence was ensured by iterating the sequence of three phases thrice ($N_{\textrm tc} = 3$). 
The selected model is the one that led to the smallest network error among all errors measured with the loss $\mathcal{L}$ selected for training during all third phase epochs in any training iterations.
All model configurations were run five times for training, and their average performance and standard deviation are reported in Table \ref{table_accuracy_Losses_Functions} of Section \ref{Comparative_Quantitative_Analysis}, which presents the impact on the results of the training of the most important hyperparameters. 


\vspace*{2mm}
\textbf{Projection of PPs: }  
In the second training phase (phase 2 in Fig.~\ref{fig_general_architecture}), latent features tensors $z_{w,h}^n \in \mathbb{R}^{D}$ ($D=128$) are extracted from each image $x_n$ of a training set: $z_{w,h}^n = f(x_n)_{w,h}$. 
Then, the prototypical-parts $p_{m,k}$ take the values of their  most similar convolutional patch tensor $z_{h,w}^n$ from the training images. 
To do so, distances $d_{z_{w,h} , p_{m,k}}$ (see Eq.~(\ref{eq_distance})) are determined between all patches $(h,w)$ of the training images $x_n$ of class $k$ and all $M$ PPs $p_{m,k}$ of same class $k$. 
Each prototype $p_{m,k}$ of class $k$ is associated with the latent feature tensor $z_{w,h}^n$ of the image patch leading to the smallest distance $d_{z_{w,h}, p_{m,k}}$ to allow for a faithful representation of the learned prototypes. 
Besides the information used by the DL model to generate final class labels, the learned prototypes are exploited to generate visual explanations.
The following update is performed when searching the projection prototype $p_{m,k}$ of class $k$, which minimizes its distance with $z_{h,w}^n$.
\begin{equation}
p_{m,k} \xleftarrow{} \operatorname*{arg\,min}_{ f(x_{n}^k) \, \in \, \mathbf{f}_{\textrm{train}} } \lVert  p_{m,k}- z_{w,h}^n \rVert_{2}  \label{eq:PP_dist} 
\end{equation}
%
%
In Eq.~(\ref{eq:PP_dist}), the set $\mathbf{f}_{\textrm{train}}$ gathers all tensors $z_{w,h}^n = f(x_{n}^k)_{w,h}$ corres\-pon\-ding to patches extracted from images $x_{n}^k$, all of which belong to class $k$. \\

\textbf{Global descriptors: } The importance of a visual perturbation $i \in\{\text{S}, \text{H}, \text{T}, \text{B}\}$ for a given PP $p_{m,k}$ is referred to as ``global descriptor'' and is determined using a set $\mathbf{X}_{\textrm{train}}$ of training images. The local scores $\Phi_{local}^{i,m,k,n}$ in the $|\mathbf{X}_{\textrm{train}}|$ training images ${x_n}$ (with ${x_n} \in \mathbf{X}_{\textrm{train}}$ and $n\in [1,|\mathbf{X}_{\textrm{train}}|]$) are weighted by the similarity scores $s_{m,k}$ (see Eq.~(\ref{Local_descriptors})) from them they are derived to obtain the global descriptor values:
\begin{equation}\label{Global_descriptors}
    \Phi_{global}^{i,m,k} = \frac{\displaystyle\sum_{n=1}^{n=|\mathbf{X}_{\textrm{train}}|} \Phi_{local}^{i,m,k,n} \cdot s_{m,k} }{\displaystyle\sum_{n=1}^{n=|\mathbf{X}_{\textrm{train}}|} s_{m,k}} 
\end{equation}
%
In contrast, if PP $p_{m,k}$ is clearly present in image $x_n$, the model will assign the prototype with a high similarity score $s_{m,k}$, which is modulated by the local score $\Phi_{local}^{i,m,k,n}$ value depending on the importance of perturbation $i$. 
In this way, PPs associated with a large similarity  $s_{m,k}$ have the highest contribution to the global descriptor value $\Phi_{global}^{i,m,k}$ for the most important local descriptors $i$ (i.e, with a large $\Phi_{local}^{i,m,k,n}$-value). Equations (\ref{Local_descriptors}) and (\ref{Global_descriptors}) provide complementary information.      
%

\begin{algorithm}[t]
\DontPrintSemicolon
\caption{Training algorithm. It is recalled that CNN $f$ consist of backbone $w_{base}$ and two $1\times 1$ convolution layers $w_{add}$. Comment start: ``//''}\label{Training_algorithm}
\KwIn{
 Pre-trained feature extractor weights $w_{pt}$,  number $N_{\text{tc}}$ of training cycles, number $N_{\text{f}}$ of training epochs for CNN $f$, number $N_{\text{h}}$  of training epochs for the FC-layer, loss $\mathcal{L}$ and learning rate $\eta$.}
\KwOut{Trained model weights and PPs $p_{m,k}$}
\textbf{Initialization:}\; 
    
    $w_{\text{base}} \leftarrow$ pre-trained weights $w_{pt}$\;
    
    $w_{\text{add}} \leftarrow$ Kaiming initialization (see \cite{KaimingInitialization_2015})\;
    
    \For{ each prototype $p_{m,k}$}{
        // RUTCI = random uniform tensor component initialization in range [0, 1]\; 
        \(p_{m,k} \leftarrow \text{RUTCI}\left([0,1]^{1 \times 1 \times D=128}\right)\)\;
    }    
    \For{ each weight $w_h \in$  ${FC}$ layer}{
    \If{$p_{m,k} \in P_k$}{$w_h^{(m,k),k} \leftarrow 1$ } 
    \Else{$w_h^{(m,k),k} \leftarrow 0$} 
    }
\textbf{Training:}\;
    
\For{$n = 1$ \KwTo $N_{\text{tc}}$}{
    \textbf{// Phase 1: Training of the CNN layers of $f$}\;
    \For{t = 1 to $N_{\text{f}}$}{
        \For{ each batch $[X, Y] \subset [X_{train}, Y_{train}]$}{
            \If{$t > 5$}{
                $w_{\text{base}} \leftarrow w_{\text{base}} - \eta \nabla_{w_{\text{base}}} \mathcal{L}(X, Y)$\;
                
                $w_{\text{add}} \leftarrow w_{\text{add}} - \eta \nabla_{w_{\text{add}}} \mathcal{L}(X, Y)$\; 
            }
        }
    }
    \textbf{// Phase 2: PPs assignment}\;
    \For{ each prototype $p_{m,k}$}{ 
        Use Eq.~(\ref{eq:PP_dist}) to update $p_{m,k}$ by finding the closest matching latent feature tensor $z_{h,w}^n$ extracted by CNN $f$ from the training images $x_n^k$ of class $k$ to which $p_{m,k}$ belongs.\;
    }
    \textbf{// Phase 3: FC-layer optimization}\;
    \For{$t = 1$ \KwTo $N_{\text{h}}$ }{
        \For{ each batch $[X, Y]$ from $[X, Y]$}{
            $w_h \leftarrow w_h - \eta \nabla_{w_h} \mathcal{L}(X, Y)$\;
        }
    }
}
\end{algorithm}

On the one hand, global score $\Phi_{global}^{i,m,k}$ highlights the significance of visual feature $i$ for PP $p_{m,k}$ without favoring a particular image. 
On the other hand, local descriptor score $\Phi_{local}^{i,m,k}$ reflects the importance of feature $i$ when assessing the similarity score $s_{m,k}$ for a given image and prototype.
The method used to generate perturbations to obtain local and global descriptors is described in (\cite{PP_Descriptors}).

In this contribution, the chosen visual features $i$ relate to texture and color information in the HSI space (hue, saturation, and intensity) since such interpretable features are also used by biologists during the MCA procedure for kidney stone type identification, which helps to make a bridge between human and DL decisions.

\subsection{Limitations of the ProtoPNet model \label{Limitations_of_PPs_models}}
The ProtoPNet model is for several reasons an inte\-res\-ting baseline for identifying kidney stones. By design, the model relies on the generated explanations to provide class labels. As sketched in Fig.~\ref{fig_general_architecture}, ``explanations'' (maps giving the similarity of PPs and input image patches) are generated before the classification of image $x_n$, which ensures consistency between model explanations and decisions. Additionally, the number of PPs for each class can be chosen to simplify the understanding of the model. 

Despite these advantages, the ProtoPNet model can also extract PP's from 
only one or few training images (even if numerous training images are available for a class), leading thus to a lack of diversity of the information included in the PP's and significantly affecting the interpretability of the decisions. This effect, referred to as PPs collapse, was demonstrated in (\cite{Flores_Araiza_2023_CVPR}) for the kidney stone identification task.
The quality of the PPs can also suffer from a lack of diversity when the training images do not carry enough discriminating information (\cite{Budach2022TheEO}).   
Another shortcoming of ProtoPNet is that the interpretability improvements reached by using PPs are sometimes obtained at the expense of accuracy, which becomes lower than that of equivalent but less interpretable DL models. It is confirmed in (\cite{ProtoPnet}) that ProtoPNet, in some cases, exhibits a lower accuracy compared to its non-interpretable CNN counterparts. 
The authors in (\cite{ProtoPnet}) used an architecture based on an ensemble approach instead of a single DL model to increase the accuracy reached by ProtoPNet architecture.
However, such an ensemble approach obscures the global rationale of the complete DL architecture since different models may focus on different parts of the input image to explain the classification or indicate visually different explanations for the same input image area. 

The main aim of this contribution is to broaden the applicability and reliability of PP-based models. Alternatives for training case-based models in a more effective way are explored to this end. In particular, new loss functions that operate in the embedding space of representative learned PPs are explored. 
Achieving such an embedding space would enhance the model generalization capabilities, improve its discriminating power, provide clearer and more interpretable results, and, in some cases, achieve scalability and computational efficiency (\cite{mendez-ruiz2023susana, gonzalez2022guided}). 

\subsection{Strategy for avoiding PPs collapse\label{Avoiding_PPs_collapse}}
%
%
Exploring novel loss functions to learn PPs is critical for enhancing the original ProtoPNet implementation by avoiding PPs collapse.  
The loss used by ProtoPNet (see Eqs.~(\ref{eq_ProtoPNet_loss}) to (\ref{eq_ProtoPNet_Sep_loss})) aims to distinguish PPs belonging to different classes and to gather PPs of same classes. 
However, this loss is not designed to prevent PPs of the same class from forming too-compact point clusters of tensors in the latent space to classify the convolutional tensor patches $z_{h,w}$ obtained from input images $x_n$ with CNN $f$. According to the distribution of the features extracted from the input images, PPs may even be learned around a single point in the latent tensor space. 
Such situations increase the risk of learning  PPs that are too similar.

Integrating Deep Metric Learning (DML) into the loss function is a strategy which can improve intra-class similarity (clusterization) and inter-class diversity (separation). 
Adequate clusterization and separation are beneficial as they help the model to recognize and reinforce the discriminating features of each class. It makes the model more capable of identifying what makes each class unique, enhancing its ability to accurately categorize new input data. Clustering is particularly important in generating case-based reasoning explanations because the latter reflects the characteristics of a class cluster that lead to the classification of the input in that cluster. 
Moreover, maintaining some separation within a particular cluster based on the main visual characteristics allows for explanations using a diverse and informative set of cases.
An appropriate balance between clusterization and separation is a key factor for designing DL models, leading to more accurate classification and precise explanations.  

\subsubsection{Existing losses and their limitation \label{Previous_losses}}

In this subsection we discuss the existing losses for PP-based models and their shortcomings

\textbf{Cross-Entropy (CE) Loss:\label{CE_loss_subsection}} This loss allows for the model to assign probabilities to each class prediction. Such an approach is advantageous in various applications (e.g., for recommendation systems), where grasping uncertainty is as crucial as achieving precise predictions. Consequently, this loss function does not impose a particular training signal for learning the PPs, but rather focuses on matching the outputs of the model to the ground truth label of its input. In Eq.~(\ref{eq_CE_loss}) defining CE-Loss $\mathcal{L}_{CE}$, \(\left|\mathbf{X}_{\textrm{train}}\right|\) stands for the number of available samples $x_n$ in learning set $\mathbf{X}_{\textrm{train}}$, $K$ is the number of classes $k$, $y_{n,k}$  is a binary indicator whose value equals 1 when label $k$ refers to the correct class for observation $x_n$, and $\hat{y}_{n,k}$ is the predicted probability (in range [0, 1]) that observation $x_n$ belongs to class $k$. 
\begin{equation}\label{eq_CE_loss}
\mathcal{L}_{CE}=-\frac{1}{\left|\mathbf{X}_{\textrm{train}}\right|}\sum_{n=1}^{\left|\mathbf{X}_{\textrm{train}}\right|}\sum_{k=1}^{K}y_{n,k}\log(\hat{y}_{n,k}) 
\end{equation}
\textbf{ProtoPNet Loss:\label{ProtoPNet_loss_subsection}}
The ProtoPNet model is designed to learn a latent space that allows for effective clustering of the most significant patches of an input image. These patches are grouped around prototypes that are semantically similar and belong to the true classes of the images. Consequently, the centers of the prototype groups (representing each a class) are distinctly separated from each other (i.e., the distance between center pairs corresponds to large $L$2-norm values).
This class separation is obtained with loss function $\mathcal{L}_{ProtoPNet}$ given in Eq.~(\ref{eq_ProtoPNet_loss}) and whose components are explained below. 
\begin{equation}\label{eq_ProtoPNet_loss}
\mathcal{L}_{ProtoPNet}= \mathcal{L}_{CE} + \mathcal{L}_{Cls} + \mathcal{L}_{Sep} + {\mathcal{L}1}
\end{equation}
While the \textit{CE-loss} $\mathcal{L}_{CE}$ penalizes misclassified samples, the minimization of the ``cluster cost'' $\mathcal{L}_{Cls}$ in Eq.~(\ref{eq_ProtoPNet_Cls_loss}) encourages each of the \(\left|\mathbf{X}_{\textrm{train}}\right|\) images $x_n$ from training set $\mathbf{X}_{\textrm{train}}$ to have at least one latent patch $z_{h,w}$ close to at least one prototype $p_{m,k}$ of its own class $k$. $P_{k}$ refers to the group of $M$ prototypes belonging to the class of image $x_n$. 
\begin{equation}\label{eq_ProtoPNet_Cls_loss}
\mathcal{L}_{Cls}= \frac{1}{\left|\mathbf{X}_{\textrm{train}} \right|} \sum_{n=1}^{\left|\mathbf{X}_{\textrm{train}}\right|}  \min_{z_{w,h},\,p_{m,k} \in P_{k} } {\| p_{m,k} - z_{w,h}\|}^2_2
\end{equation}
The minimization of \textit{separation cost} $\mathcal{L}_{Sep}$ given in  Eq.~(\ref{eq_ProtoPNet_Sep_loss}) favors a latent patch $z_{h,w}$ of a training image $x_n$ to stay far away from the prototypes $p_{m,k}$ which do not belong to its own class $k$, which is mathematically formulated by $p_{m,k} \notin P_{k}$.
\begin{equation}\label{eq_ProtoPNet_Sep_loss}
\mathcal{L}_{Sep}= -\frac{1}{\left|\mathbf{X}_{\textrm{train}} \right|} \sum_{n=1}^{\left|\mathbf{X}_{\textrm{train}}\right|}  \min_{ z_{w,h},\, p_{m,k} \not \in P_{k} } {\| p_{m,k} - z_{w,h}\|}^2_2
\end{equation}
In Eq.~(\ref{eq_ProtoPNet_loss}), $\mathcal{L}1$ is a regularization term that prevents the FC-layer of the model from learning excessively large weights. $\mathcal{L}1$ stands for the 
$L1$-norm of the weight parameters of the FC layer.Using this norm encourages sparsity of the weights of the model, of feature extractor $f$ consisting of $w_{base}$ and $w_{add}$ (see beginning of Section \ref{Components_of_inference} and Algorithm \ref{Training_algorithm}), and of  the last FC-layer $w_h$.

The four components in Eq.~(\ref{eq_ProtoPNet_loss}) shape the latent space into a semantically meaningful cluster structure, facilitating the $L2$ distance-based classification of the ProtoPNet network.  
However, the richness of the variety of the training data can be lost for some classes without a term preventing the collapse of the learned PPs.
\subsubsection{Proposed losses \label{Proposed_losses}}
The effectiveness of a loss function that only impacts the behavior of the PPs-layer during training is explored in this section.
This implies that the objective function does not affect the FC-layer that provides the final class labels. \newline

\textbf{ICNN loss: \label{ICNN_loss_subsection}}
The Inter- and Intra-Class Nearest Neighbor Score (ICNN-Score) is a latent feature space score that can be used in a loss 
%

As expressed in Eq. (\ref{eq_ICNN}), this score (to be maximized) results from the product of three functions $\Lambda$, $\Omega$, and $\Gamma$. 
\begin{align} 
\label{eq_ICNN}
\hspace*{-10mm}\mathrm{ICNN} &=\\ & \hspace*{-2mm} \frac{1}{\left|  \mathbf{X}_{\textrm{train}} \right|} \hspace*{-2mm} \sum_{n=1}^{\left|\mathbf{X}_{\textrm{train}}\right|} \hspace*{-2mm} \Lambda\left(z_{w,h},p_{m,k}\right) \Omega\left(z_{w,h},p_{m,k}\right) \Gamma\left(z_{w,h},p_{m,k}\right)  \nonumber
\end{align}

As detailed in Appendix \ref{Appendix_A}, the ICNN score used in this contribution was adapted from the method in (\cite{mendez-ruiz2023susana}). This approach gives an accurate estimate of the ICCN score, even in scenarios with few available points (PPs) in the latent feature space. 

To sum up, functions $\Lambda$, $\Omega$ and $\Gamma$ have following roles in the ICNN score given in Eq.~\eqref{eq_ICNN}. 
Function $\Lambda$ allows to select PPs that help to represent classes by compact clusters while maximizing the inter-cluster distances in the latent feature space.
%
Function $\Omega$ (which modulates the values of function $\Lambda$)
helps to select PPs able to maintain diversity and representativity even in a compact classes.    
The $\Gamma$ function increases the classification accuracy by selecting PPs class configurations helping the latent feature tensors  $z_{w,h}^k$ of class $k=i$ to be close to a high number of PPs $p_{m,k}$ of class $k=i$ while being near to few PPs of other classes ($k\neq i$). 
The design of suitable functions $\Lambda$, $\Omega$ and $\Gamma$ for the kidney stone identification task is detailled in Appendix \ref{Appendix_A}. 

The Loss $\mathcal{L}_{\mathrm{ICNN}}$ is determined with score $ICNN$ as follows.
\begin{equation}\label{eq_ICNN_loss}
    \mathcal{L}_{\mathrm{ICNN}} = -\log\left(\mathrm{ICNN}\left(x_n\right)\right)
\end{equation}
In Eq.~(\ref{eq_ICNN_loss}), the value of $\mathcal{L}_{\mathrm{ICNN}}$ tends towards 0 (a high value) when score $ICNN$ is close to 1. \newline
\textbf{CIC Loss:\label{CIC_loss}}
The CE loss$\mathcal{L}_{CE}$ and the ICNN-Loss were combined in loss $\mathcal{L}_{CIC}$ (see Eq.~(\ref{eq_CIC_loss})) to train a self-explainable architecture. 

The CE loss component $\mathcal{L}_{CE}$ is responsible for optimizing the performance of the model by fine-tuning the parameters of the model, striving for accurate predictions.
Meanwhile, the ICNN loss $\mathcal{L}_{CIC}$ component refines the parameters of the feature extractor $f$ by enhancing the clustering of different image classes in the latent feature space. 

This improved clusterization is intended to facilitate the learning of Prototypical Parts (PPs) for each class with a more diverse and representative set of characteristics for their respective classes. 
Consequently, the model not only achieves higher accuracy but also learns richer and more interpretable PPs, which are essential for explaining the classification decisions in the presented case-based reasoning framework.

\begin{equation}\label{eq_CIC_loss}
    \mathcal{L}_{CIC} = \mathcal{L}_{CE} + \mathcal{L}_{\mathrm{ICNN}} 
\end{equation}
\textbf{PPIC Loss:\label{ProtoPNet_plus_ICNN_loss}}
The PPIC Loss combines the ProtoPNet loss function with the ICNN Loss (see loss $\mathcal{L}_{PPIC}$ in Eq.~(\ref{eq_ProtoPNet_plus_ICNN_Loss})) to train the architecture. The joint use of these two losses, which are aimed at organizing the latent space into a semantically meaningful clustering structure,  helps to investigate whether this loss association can synergistically enhance the clustering or not and increase the model’s accuracy. 
\begin{equation}\label{eq_ProtoPNet_plus_ICNN_Loss}
    \mathcal{L}_{PPIC} = \mathcal{L}_{ProtoPNet} + \mathcal{L}_{\mathrm{ICNN}} 
\end{equation}
%
\begin{figure*}[t] 
    \centering
    \subfloat[\textbf{SUR}]{
    \label{fig_dataset_a}\includegraphics[width=0.50\linewidth]{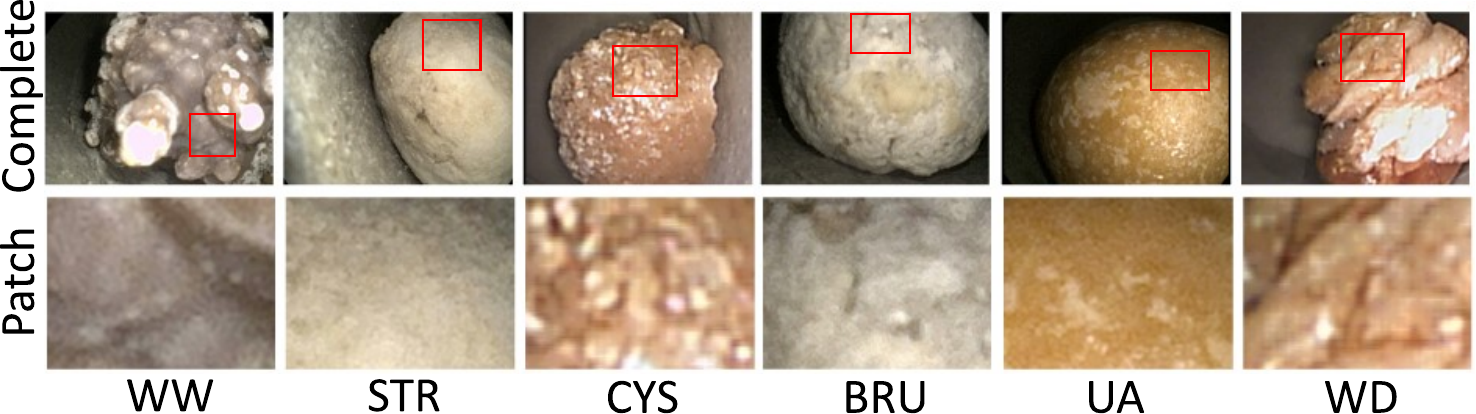}} 
    \subfloat[\textbf{SEC}]{
    \label{fig_dataset_b}\includegraphics[width=0.50\linewidth]{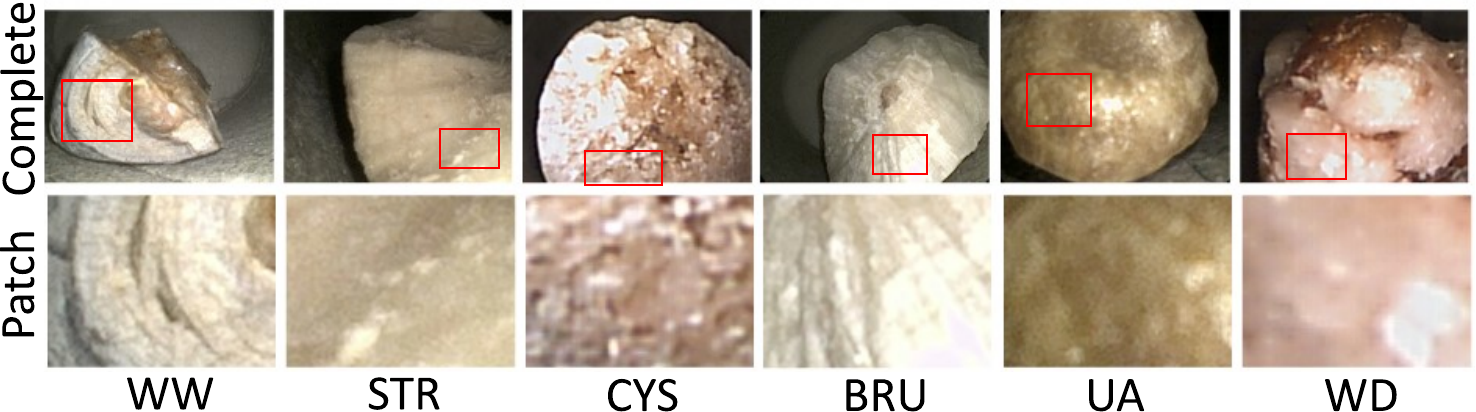}}
    \vspace{-0.1cm}
    \caption{Examples of the six most common kidney stone types: Whewellite (WW), Struvite (STR), Cystine (CYS), Brushite (BRU), Uric Acid (UA), and  Weddellite (WD). Surface (SUR) and section (SEC) views of the kidney stone fragments are given in figures (a) and (b), respectively. The complete kidney stone images are given in the upper rows of the figures, while the lower rows represent patches extracted from the images in the upper rows.  
    }
    \label{fig_dataset}
\end{figure*}
\section{Experimental Setup\label{Experimental_Setup}}   
Various model configurations
(i.e., with different feature extraction backbones, loss functions, number of PPs per class) were implemented on two P100 GPUs controlled by a Nvidia DGX1 server, each GPU having 16 GB of VRAM.  The code was compiled with the Nvidia Cuda Compiler (NVCC) version V11.6.112 and written with Python (version 3.8.12) with imports from the  Pytorch library (version 1.12).
Except for the data augmentation, which is described in Section \ref{Preproces_and_Transfer_learning}, the complete initialization and hyperparameter training phases are those detailed in (\cite{ProtoPnet}).
 Experiments were carried out i) to explore the effectiveness of different backbones used as feature extractor network $f(x_n)$, ii) to find the optimal number of PPs per class, and iii) to assess the importance of data augmentation. These tests allow to evaluate the performance of the model and the accuracy of the explanations obtained for different architecture configurations.
    \begin{table}[]
\centering
\resizebox{\columnwidth}{!}{%
\begin{tabular}{@{}cccccc@{}}
\toprule
Subtype & Main component & Label & Surface & Section & Mixed \\ \midrule  \vspace{-0.05cm}
Ia & Whewellite & WW & 62 & 25 & 87 \\  \vspace{-0.05cm}
IIa & Weddellite & WD & 13 & 12 & 25 \\  \vspace{-0.05cm}
IIIa & Uric Acid & UA & 58 & 50 & 108 \\  \vspace{-0.05cm}
IVc & Struvite & STR & 43 & 24 & 67 \\  \vspace{-0.05cm}
IVd & Brushite & BRU & 23 & 4 & 27 \\  \vspace{-0.05cm}
Va & Cystine & CYS & 47 & 48 & 95 \\ \cmidrule(l){3-6}   \vspace{-0.05cm}
 &  & TOTAL & 246 & 163 & 409 \\ \bottomrule  \vspace{-0.05cm}
\end{tabular}
}
\caption{The acquired endoscopic images form three datasets: two data sets consist of images from a unique view (i.e., the surface and section views gathering 246 and 163 images respectively), while the third dataset referred to as ``Mixed'' consists of all 409 images of both view.}
\label{tab_dataset}
\end{table}

\subsection{Dataset}\label{Kidney_stone_dataset}
A simulated in-vivo dataset of kidney stone images was used for the experiments reported in this contribution. Kidney stone fragments extracted from patients were successively placed in a tube whose internal wall has a color close to that of the epithelium of ureters. Two different reusable Karl Storz digital ureteroscopes connected to two video card systems (Storz Image 1 Hub$^{\textrm{TM}}$ and Storz Image1 S$^{\textrm{TM}}$) were used to acquire images in the tubular shaped environment that simulates real conditions since, as during an ureteroscopy,  the illumination strongly changes with the viewpoint, the images are affected by motion blur and specular reflections, and the endoscope's viewpoint cannot be exactly controlled. Details of this acquisition protocol can be found in (\cite{elbeze2022evaluation}).  As noticeable in Table \ref{tab_dataset}, the dataset consists of 246 surface images and 163 section images, these two types of images being referred to as section and surface ``views'' in this contribution. The two views contain six of the most common kidney stone subtypes: Ia (Whewellite, WW), IIa (Weddellite, WD), IIIa (anhydrous Uric Acid, UA), IVc (Struvite, STR), IVd (Brushite, BRU), and Va (Cystine, CYS). The kidney stone subtypes were determined with the MCA-procedure described in Section \ref{Intro-MCA}, which provides the ground truth (i.e., the class labels).

The DL-based models were trained and tested using 12,000 square patches of 256$\times$256 pixels extracted from the 409 endoscopic images of both views. Similarly to the images of Table \ref{tab_dataset}, the whole set of 12,000 patches is referred to as \textit{mixed} dataset. Images and patches of the dataset are shown in Fig.~\ref{fig_dataset}. 
The optimal size of the patches and how redundant information can be avoided by limiting the patch overlap is discussed in (\cite{lopez2024vivo}).

The different tested DL models were trained with the method described in Section \ref{Training_procedure} using only the mixed dataset described in Table \ref{tab_dataset}. 
The decision to only train the models on the dataset with mixed views (without using the datasets including only one type of view, as in numerous other contributions (\cite{black2020deep, estrade2022towards, black2020deep, lopez2021assessing, lopeztiro23boosting, villalvazo2023improved})) lies on the willingness to be in accordance with the MCA-procedure in which the biologists simultaneously use surface and section view information to visually perform the morphological analysis of kidney stone fragments (\cite{villalvazo2023improved}). 
%

\subsection{Pre-processing and transfer learning \label{Preproces_and_Transfer_learning}} 
Train and test sets included  80\% of the patch-set (i.e., 1,600 images per class) and 20\% of the patch-set (i.e., 400 images per class), respectively. 
The patches were also ``whitened" using mean $m_{i}$ and standard deviation $\sigma_{i}$ of color values $I_{i}$ in each channel $(I^{w}_{i} = (I_{i}-m_{i})/\sigma_{i}$, with $i=R,G,B)$. 

The DL models were trained for all losses described in Sections \ref{Previous_losses} and \ref{Proposed_losses}. 
The different training configurations (as detailed later, with different feature extraction backbones, various numbers of PPS, CNN-models with and without PPs, etc.)  were performed on the mixed view data described in Section \ref{Kidney_stone_dataset}, with and without data augmentation. 
In the preprocessing phase, data augmentation involved the stochastic selection of one specific geometric image modification out of a set of possible transformations.
The set of transformations included random horizontal and vertical flipping, in-plane rotations ranging from -180 to 180 degrees, perspective distortions of up to 40\%, scaling changes of up to 50\%, translations of up to 20\% of the image dimensions in both vertical and horizontal directions, and symmetric padding extending to 50 pixels on all sides.
Once selected, the transformation has a 50\% chance of being applied to the image.  

After training 'i.e.,in inference mode) the different model configurations are used to obtain the PPs \textit{descriptors}, the aforementioned set of perturbations (i.e., $i$ = S, H, T or B for saturation, hue, texture, or brightness, respectively) being applied on the test images to assess and compare the model performances subjected to perturbations. 

\subsection{Model configurations}\label{Model_configurations}
Three different CNN-architectures were taken as backbone of the DL model: 
i) \textit{VGG16} is used to examine the performance of a simple deep CNN, 
ii) \textit{ResNet50} shows the efficiency of a medium-sized CNN with residual connections, and iii) \textit{DenseNet201} allows to assess a model with dense connections. 
These three architectures are in the PyTorch library and were pre-trained on ImageNet. Their CNN layers, also known as feature extractors, were used as the backbone of models trained (i.e. fine-tuned) on the mixed kidney stone dataset. 
These three CNNs were also chosen since several works used them to identify the type of kidney stones (\cite{martinez2020towards, estrade2022towards, black2020deep, lopez2021assessing, villalvazo2023improved}). Thus, these networks are appropriate for a baseline comparison. 

The impact on the class labels of the variations of an important hyperparameter in the proposed approach (i.e., the number of PPs in the prototype layer)  was also investigated. Model configurations were tested for the following PPs numbers: 1, 3, 10, 50, and 100 PPs were used for each of the six classes and each CNN-backbone. 

One of the most important design decisions in the proposed model is related with the choice of the appropriate loss functions by assessing their ability to avoid PPs-collapse (see Section \ref{Avoiding_PPs_collapse}). The 
following loss functions were explored:
\vspace{-\topsep}
\begin{itemize}
\setlength\leftmargin{1.5em}
\setlength{\parskip}{0pt}
\setlength{\itemsep}{0pt plus 1pt}
\item \textit{Categorical Cross-Entropy Loss}: the $\mathcal{L}_{CE}$ loss given in Eq.~(\ref{eq_CE_loss}) is used with the default setting from the Pytorch library.
\item \textit{ProtoPNet Loss}: while in Eq.~(\ref{eq_ProtoPNet_loss}) $\mathcal{L}_{CE}$, $\mathcal{L}_{Cls}$, $\mathcal{L}_{Sep}$ and $\mathcal{L}1$ have an equal contribution in loss $\mathcal{L}_{ProtoPNet}$, in the performed experiments the impact of these four loss-components on the global loss was controlled by empirically chosen weights (i.e., weights of 1, 0.8, 0.08, and 1e-4 were applied to $\mathcal{L}_{CE}$, $\mathcal{L}_{Cls}$, $\mathcal{L}_{Sep}$ and $\mathcal{L}1$, respectively). 

\item\textit{CIC Loss} given in Eq.~(\ref{eq_CIC_loss}): 
the values of the ICNN score parameters (see Eq.~(\ref{eq_ICNN})) used to compute loss $\mathcal{L}_\mathrm{ICNN}$ defined in Eq.~(\ref{eq_ICNN_loss}) were set as follows: $p = q = r = 1$. Losses $\mathcal{L}_{CE}$ and $\mathcal{L}_{ICNN}$ were equally weighted for two main purposes. On the one hand, it allows to understand the DL model behavior by giving the same importance to the model accuracy (depending mainly on $\mathcal{L}_{CE}$) and to the clustering quality of the training samples within and between classes mainly relating to loss $\mathcal{L}_{ICNN}$. On the other hand, equally weighted losses $\mathcal{L}_{CE}$ and $\mathcal{L}_{ICNN}$ enable to establish a performance baseline for the $\mathcal{L}_{CIC}$ Loss (see Table \ref{table_accuracy_Losses_Functions}).

\item \textit{PPIC Loss} given by Eq.~(\ref{eq_ProtoPNet_plus_ICNN_Loss}): the ICNN score used in the $\mathcal{L}_\mathrm{ICNN}$ loss component of loss $\mathcal{L}_{PPIC}$ is again determined for $p\,$=$\,q\,$=$\,r\,$=$\,1$ in Eq.~(\ref{eq_ICNN}) and losses $\mathcal{L}_{ProtoPNet} $ and $\mathcal{L}_\mathrm{ICNN}$ were also equally weighted. 
\end{itemize}
\vspace{-\topsep}
\subsection{Model evaluations}\label{Model_Evaluations}
The performance of the trained models is assessed using the accuracy criterion, which is the most commonly used metric in the literature for kidney stone identification. In Eq.~(\ref{Acc}), $\textrm{TP}_k$ (true positives) and  $\textrm{TN}_k$ (true negatives) respectively represent the number of correctly predicted instances and the number of correctly predicted instance absences for class $k$ in the whole set of input data. On the contrary, $\textrm{FP}_k$ and $\textrm{FN}_k$ (false positives and false negatives) stand respectively for the incorrectly predicted number of instances and instance absences for class $k$.   
\begin{equation}\label{Acc}
   Acc_{k} = \frac{\textrm{TP}_k + \textrm{TN}_k }{\textrm{TP}_k  + \textrm{TN}_k  + \textrm{FP}_k  + \textrm{FN}_k }   
\end{equation}
Accuracy $Acc$ is given by the weighted average of all $Acc_{k}$ values, the weights being given by the number of instances of the classes in the ground truth.
Each model configuration was trained five times. All training sessions were performed on the mixed view dataset that contains both the surface and the section views of kidney stone fragments. Each of the five training runs had a different initialization seed. The average $\overline{Acc}$ and standard deviation $\sigma_{acc}$ of the five accuracy values $Acc$ are given to allow for a comparison of the performances obtained by the methods in the literature (see Table \ref{table_SOTA_vs_ours}) and that of the different configurations tested for the proposed model (see Table \ref{table_accuracy_Losses_Functions}). 
\begin{table}[t]
\centering
\caption{Accuracy of state-of-the-art models against the proposed DL architecture. The models were trained five times with data augmentation on the mixed view of the kidney stone data set.}
\label{table_SOTA_vs_ours}
\resizebox{\columnwidth}{!}{%
\begin{tabular}{cclc}
\hline
\textbf{Method type}    & \multicolumn{2}{c}{\textbf{Contribution}}             & $\overline{Acc}\pm\sigma_{acc}$     \\ \hline
\multirow{4}{*}{CNN based} & \multicolumn{2}{c}{Black et al. (2020)}         & 80.1±13.8          \\
                           & \multicolumn{2}{c}{Lopez et al. (2021)}         & 85.0±3          \\
                           & \multicolumn{2}{c}{Estrade et al. (2022)}       & 70.1±22.3          \\
                           & \multicolumn{2}{c}{\cite{lopeztiro23boosting}}    & 85.6±0.1        \\ \hline
\multirow{5}{*}{PPs-based} & \multicolumn{2}{c}{Chen et al. (2019a)}         & 87.3±0.9          \\
                           & \multicolumn{2}{c}{Nauta et al. (2021a)}        & 85.2±7.4          \\
                           & \multicolumn{2}{c}{Rymarczyk et al. (2022)}     & 85.6±1         \\
                           & \multicolumn{2}{c}{Flores-Araiza et al. (2023)} &88.2±2.1          \\
                           & \multicolumn{2}{c}{This contribution}                        & \textbf{90.4±0.6} \\ \hline
\end{tabular}%
}
\end{table}

\begin{table*}[b]
\centering
\resizebox{\textwidth}{!}{
\begin{tabular}{ccccccccc}
\hline
\multicolumn{2}{c}{} &
  \multicolumn{5}{c}{Prototypical-Parts based models} &
   &
  CNN models \\ \hline
\multicolumn{1}{l}{Backbone} &
  \# of PPs &
  \begin{tabular}[c]{@{}c@{}}$\overline{Acc}\pm\sigma_{acc}$\\      {\em CE} Loss\end{tabular} &
  \begin{tabular}[c]{@{}c@{}}$\overline{Acc}\pm\sigma_{acc}$\\      {\em ProtoPNet Loss}\\      ($\epsilon = 0.0001$)\end{tabular} &
  \begin{tabular}[c]{@{}c@{}}$\overline{Acc}\pm\sigma_{acc}$ \\      {\em ProtoPNet Loss}\\      ($\epsilon = 0.01$)\end{tabular} &
  \begin{tabular}[c]{@{}c@{}}$\overline{Acc}\pm\sigma_{acc}$\\      {\em CIC Loss} \\      \end{tabular} &
  \begin{tabular}[c]{@{}c@{}}$\overline{Acc}\pm\sigma_{acc}$\\      {\em PPIC Loss} \end{tabular} &
   &
  \begin{tabular}[c]{@{}c@{}}$\overline{Acc}\pm\sigma_{acc}$\\      CE Loss\end{tabular} \\ \hline
\multirow{5}{*}{VGG16} &
  1 &
  81.71±1.88 &
  81.78±1.60 &
  84.27±0.90 &
  82.59±1.55 &
  84.01±2.18 &
   &
  \multirow{5}{*}{81.47±2.72} \\
 &
  3 &
  82.75±2.78 &
  82.21±3.33 &
  83.91±1.44 &
  83.72±1.54 &
  83.86±1.50 &
   &
   \\
 &
  10 &
  83.72±2.57 &
  82.08±0.90 &
  81.98±3.49 &
  82.98±1.74 &
  82.55±1.91 &
   &
   \\
 &
  50 &
  82.06±1.47 &
  82.61±1.95 &
  82.39±1.48 &
  83.27±2.72 &
  80.79±1.19 &
  \textbf{} &
   \\
 &
  100 &
  84.18±1.77 &
  79.70±3.20 &
  82.50±2.69 &
  82.22±1.54 &
  82.13±3.79 &
   &
   \\ \hline
\multirow{5}{*}{ResNet50} &
  1 &
  87.42±1.83 &
  \textbf{88.21±2.07} &
  86.48±2.34 &
  87.40±1.72 &
  89.57±0.48 &
   &
  \multirow{5}{*}{83.40±5.32} \\
 &
  3 &
  89.65±2.11 &
  86.66±1.37 &
  87.67±1.54 &
  \textbf{90.37±0.58} &
  \textbf{90.30±1.84} &
  \textbf{} &
   \\
 &
  10 &
  89.50±0.78 &
  85.44±1.44 &
  86.40±1.31 &
  89.98±1.09 &
  89.24±1.07 &
  \textbf{} &
   \\
 &
  50 &
  89.12±0.85 &
  85.25±2.15 &
  88.28±1.50 &
  90.32±0.80 &
  88.28±1.44 &
  \textbf{} &
   \\
 &
  100 &
  \textbf{90.02±0.86} &
  86.52±1.42 &
  86.77±3.00 &
  90.20±0.67 &
  87.66±1.53 &
  \textbf{} &
   \\ \hline
\multirow{5}{*}{DenseNet201} &
  1 &
  86.27±1.79 &
  86.29±1.91 &
  \textbf{88.36±2.80} &
  86.89±1.60 &
  90.16±1.33 &
   &
  \multirow{5}{*}{\textbf{89.67±3.60}} \\
 &
  3 &
  88.02±1.99 &
  85.19±1.50 &
  86.02±1.69 &
  86.96±2.59 &
  88.47±2.14 &
   &
   \\
 &
  10 &
  88.48±1.37 &
  87.29±0.92 &
  87.81±2.12 &
  89.84±0.72 &
  89.13±1.62 &
  \textbf{} &
   \\
 &
  50 &
  87.34±1.97 &
  85.39±1.08 &
  86.14±2.37 &
  89.07±1.01 &
  88.20±1.51 &
  \textbf{} &
   \\
 &
  100 &
  88.23±1.07 &
  83.62±3.18 &
  85.89±1.77 &
  88.53±2.80 &
  86.90±2.81 &
  \textbf{} &
   \\ \hline
\multicolumn{2}{c}{\begin{tabular}[c]{@{}c@{}}Average Accuracy per\\      training loss function\end{tabular}} &
  86.56±3.25 &
  85.02±2.83 &
  85.66±2.85 &
  86.96±3.42 &
  86.75±3.55 &
  \textbf{} &
  84.84±5.29 \\ \hline
\end{tabular}%
}
\caption{Accuracy values obtained for different configurations of the ProtoPNet architecture trained with various loss functions. The rightmost column gives the accuracy obtained with the black-box CNN used as the backbone in the different ProtoPNet configurations. All the models were trained five times over the mixed-view kidney stone dataset.\label{table_accuracy_Losses_Functions}}
\end{table*}
%
Local descriptors were calculated (as presented in Section \ref{Components_of_inference}) for all the models (with different backbones), with the losses specified in Section \ref{Model_configurations}, i.e., $\mathcal{L}_{CE}$, $\mathcal{L}_{ProtoPNet}$, $\mathcal{L}_{CIC}$ and $\mathcal{L}_{PPIC}$. These local descriptors, determined for the six subtypes of kidney stones, allow to identify the relevant visual characteristics learned by the PPs. 

A T-SNE dimensionality reduction technique (\cite{tSNE}) is used in the latent feature space.
This technique is only applied to the most accurate models and allows for a qualitative evaluation of the clustering properties of their latent space. This comparison of clustering obtained for different models seeks to check whether the greater accuracy of a model is effectively correlated with a more effective clustering.

Frechet Inception Distance (FID) scores are also reported in Section \ref{Comparative_Quantitative_Analysis} to provide a quantitative measure of the similarity between the distribution of the learned PPs and that of the training dataset. 
Moreover, the performances of the models with the best accuracy were quantitatively assessed for each class by comparing i) the mean, variance, and standard deviation of the distances between all pairs of features extracted from the training input samples and the PPs with ii) the three same statistical values of the distances between all pairs of the learned PPs. This comparison illustrates the clustering characteristic distances obtained in the best scenario by each of the different losses described in Sections \ref{Previous_losses} and \ref{Proposed_losses}.

\section{Results \label{Results}}
The average accuracy percentages obtained for different configurations of the model described in Section \ref{Proposed_Framework} are presented in Table \ref{table_accuracy_Losses_Functions}.

\subsection{Quantitative analysis \label{Comparative_Quantitative_Analysis}}

One of the most important aims of any XAI method is to provide explainability capabilities while reaching performances (e.g., for instance, accuracy) similar to that of traditional DL methods.  
Table \ref{table_SOTA_vs_ours} shows that the most accurate XAI model tested in this contribution outperforms the results obtained in the state-of-the-art. The best model configuration led to an average accuracy of 90.37\%, which is 4.7\% better than the best method using a CNN-method (\cite{lopeztiro23boosting}). It is noticeable that both the XAI and CNN-methods used the \textit{ResNet50} backbone. 
Moreover, the proposed XAI-arhitecture achieved an accuracy improvement of 2.1\% over the best XAI method dedicated to the identification of kidney stones (\cite{Flores_Araiza_2023_CVPR}) using mixed views.

Table \ref{table_accuracy_Losses_Functions} gives an overview of the average accuracy for various configurations of the proposed DL model trained with different loss functions and for three feature extraction backbones. 
Additionally, the rightmost column gives the accuracy of the ``black-box'' CNNs used as backbones in the XAI models as a baseline for the performance assessment. 

The highest average accuracies were obtained with the Resnet50 backbone associated with i) the \textit{CE + ICNN Loss} function and ii) the \textit{ProtoPnet + ICNN loss} (values in bold in the sixth and seventh columns of Table \ref{table_accuracy_Losses_Functions}). 
Furthermore, it is noticeable in this table that PP-based models often outperform their pure CNN counterparts, especially when the Resnet50 backbone is used. This performance increase is also observable when both models are trained using the same loss function (i.e., the \textit{CE loss} in the third and last columns in Table \ref{table_accuracy_Losses_Functions}) due to an adequate selection of hyperparameters.   
Moreover, it can be noticed in Table \ref{table_accuracy_Losses_Functions} that PP-based models trained with the \textit{CE Loss} often outperform the same PP-models trained with the \textit{ProtoPNet loss} function.  
For the proposed ProtoPNet architecture, a loss function solely focusing on a performance metric (such as \textit{CE loss} does) can effectively outperform the performance obtained with the \textit{original ProtoPNet loss}, which considers the properties of the trained PPs used to explain the classifications. 

Concerning the CNN-backbones used by the PP-based models, those with residual connections led clearly to a higher accuracy than those without such connections. Thus, the training configurations using \textit{ResNet50} and \textit{DenseNet201} achieved respectively a mean accuracy of 88.27\% and 87.38\%, while \textit{VGG16} led to a mean accuracy of 82.64\%. The configurations implementing a \textit{ResNet50} exhibit a mean accuracy improvement of 5.63\% and 0.89\% over the \textit{VGG16} and \textit{DenseNet201} configurations, respectively. 

It is noticeable that assigning a higher number of PPs per class does not present a considerable or systematic improvement in the performance of PPs-based models for the targeted classification task. 

It is interesting to notice in Table \ref{table_accuracy_Losses_Functions} that all PPs-based models with a \textit{ResNet50} backbone trained with a loss function integrating the $\mathcal{L}_{\textrm{ICNN}}$ loss clearly outperform their purely CNN counterpart (i.e., the baseline \textit{ResNet50} network) in terms of accuracy. 

\begin{figure*}[ht] 
    \centering
    \includegraphics[width=0.80\linewidth]{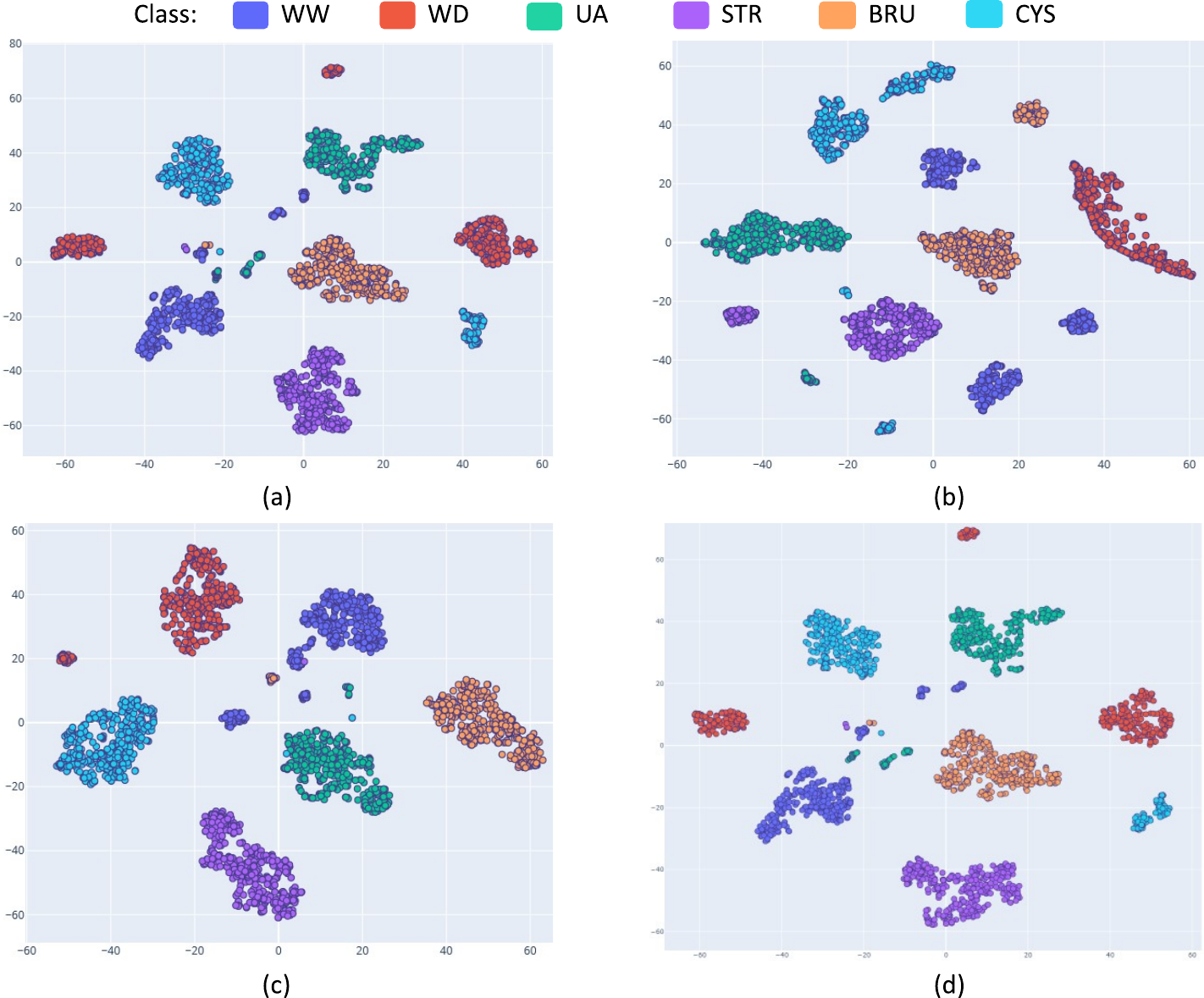}
    \caption{
    \small{
    \textbf{Qualitative clusterization comparison according to the loss functions.} 
    The sub-figures show each a 2D t-SNE plot of the latent space of the $z_{w,h}$ tensors encoded with a ResNet50 backbone, trained with three PPs per class, and a specific loss function, namely {\em CE Loss} function $\mathcal{L}_{CE}$ in (a), {\em ProtoPNet loss} function $\mathcal{L}_{ProtoPNet}$ in (b), {\em CIC loss} function $\mathcal{L}_{CIC}$ in (c), and {\em PPIC Loss} function $\mathcal{L}_{PPIC}$ in (d). 
    }
    } 
    \label{fig_TSNE_comparison}
\end{figure*}

\vspace*{2mm}
 The similarity between the images that include PPs and the training distribution was evaluated to guarantee that the learned PPs are as much as possible representative of the image distribution in the training dataset. This representability is crucial to fully exploit a diverse and unbiased description of the training dataset in the PPs and using it for inference explanations.
The \textit{Frechet Inception Distance (FID)} serves as a proxy metric of the desired property of the learned PPs. This measure provides scores that quantify the similarity between two image distributions in a common embedding space of an \textit{InceptionV3} model trained in ImageNet. 
In this context, a lower FID score is a measure of high similarity. 
The FID-score was assessed between the subset of training images learned to be the PPs over the five training runs of models with a specific loss function and one hundred PPs per class and the dataset, training, and test sets.
In this contribution, the models trained with the \textit{CIC loss} achieved the lowest FID score of 47.26. In comparison, the FID score obtained by the models trained with the original \textit{ProtoPNet loss} had an FID score of 59.51.  
Thus, the \textit{CIC loss} allows to learn the most similar PPs (i.e., the closest visual cases) that best represent the training data distribution.

Then, it was determined which loss function produced the PPs images with a \textit{closer} representation of new inference cases belonging to the test set.
The minimum expected FID score measured between the training and test sets was calculated as a baseline,  
The obtained an FID score of 44.78 represents the natural distance between the training images and the distribution of inference images of  the test set.
The test set and the PPs images from the models trained with \textit{CE + ICNN loss} led to an FID score of 75.88 (one hundred PPs per class were used to obtain this result). 
Similarly, a FID score of 88.12 was obtained using the test set and PPs images of models trained with the \textit{ProtoPNet loss}. This result was again obtained for one hundred PPs per class.
These measurements position the PPs images from models trained with the \textit{CIC loss} as the most similar PPs between the models trained under different loss functions to new images during inference on the test set.
 
\subsection{Qualitative Analysis\label{Qualitative_Analysis}}
\vspace{1mm}
{\bf Representations with the t-distributed Stochastic Nei\-ghbor Embedding (t-SNE) algorithm.}
Figure \ref{fig_TSNE_comparison} gives the 2D t-SNE plots of the latent space learned by the models integrating the \textit{ResNet50} backbone, exploiting three PPs per class and using one among the four loss functions under study.
Each point in this 2D-space represents the most similar, and therefore closest, feature patch tensor $z_{w,h}$ from each test image to a learned PPs. 
For six classes (i.e., six kidney stone types), six clusters of grouped points, with the smallest possible extend, and without outliers (groups of few isolated points) should ideally be visible in these plots. Thus, these plots allow for a qualitative (visual) appreciation of the clusterization of each training loss function.

It can be noticed in Figs.~\ref{fig_TSNE_comparison}.(a) and \ref{fig_TSNE_comparison}.(b)  that the 2D t-SNE spaces of the two loss functions ($\mathcal{L}_{CE}$ and $\mathcal{L}_{ProtoPNet}$) that do not integrate the ICCN-score  produce clusters split in sub-clusters which may be separated by clusters of other classes. 
For instance, in the 2D t-SNE space obtained for the \textit{CE loss} function (see Fig.~\ref{fig_TSNE_comparison}.(a)), the cystine class (CYS, light blue points) mainly consists of two sub-clusters separated by the brushite class (BRU, orange points) and outliers from other classes. 
A similar observation can be made for the 2D t-SNE space obtained for the \textit{ProtoPNet loss} (see Fig.~\ref{fig_TSNE_comparison}.(b)): three whewellite  (WW, dark blue points) sub-clusters surround the largest of the two BRU class sub-clusters. 

This class splitting and separation issues are strongly mitigated when the explored model makes use of a loss function including the $\mathcal{L}_\textrm{ICNN}$ loss. As shown in Fig.~\ref{fig_TSNE_comparison}.(c), the \textit{CIC loss} ($\mathcal{L}_{PICCN}= \mathcal{L}_{CE}+\mathcal{L}_\textrm{ICNN}$) led to six classes represented by one main (large) sub-cluster completed by few small sub-clusters or outliers, with a very moderate sub-class separation. Such clustering is needed when models have to generalize well since new samples of different classes will naturally be more separated and easier to classify. 

An unexpected result is that the model trained with the \textit{PPIC loss} function $\mathcal{L}_{PPIC}= \mathcal{L}_{ProtoPNet}+\mathcal{L}_\textrm{ICNN}$ (see Fig.~\ref{fig_TSNE_comparison}.(d)) led to two large and separated sub-clusters for the weddellite (WD, red points) and cystine (CYS, sky-blue points) classes. This cluster configuration in the 2D t-SNE space is most similar to that of the model trained solely with the \textit{CE loss} function $\mathcal{L}_{CE}$ (see Fig.~\ref{fig_TSNE_comparison}.(a)), which indicates a countering effect between some of the loss terms in the $\mathcal{L}_{ProtoPNet}$ loss and the $\mathcal{L}_\textrm{ICNN}$ loss that form the $\mathcal{L}_{PPIC}$ loss function, leaving out just the effect of the $\mathcal{L}_{CE}$ loss from the $\mathcal{L}_{ProtoPNet}$ loss.

The clustering quality of the latent feature tensors $z_{w,h}$ was also assessed on the data used to generate the 2D t-SNE plots (in Fig.~\ref{fig_TSNE_comparison}). This is, the closest $z_{w,h}$ tensor, per test image, to the PPs, were subjected to 5-fold cross-validation using the $k$-nearest neighbor ($k$NN) technique, with $k\,$=$\,5$. The average accuracies obtained for the clusters of the six classes of kidney stones accordingly to the best models (based on \textit{ResNet50} backbone, three PPs per class) trained with the \textit{CE-}, \textit{CIC-}, \textit{ProtoPNet-}, and \textit{PPIC-}losses were 89\%, 88\%, 86\%, and 91\%, respectively, as shown in Table \ref{table_KNN}. 
The $k$NN algorithm successfully classified test samples with an accuracy comparable to that of the trained FC-layer of each model. Indeed, compared to the accuracies given in Table \ref{table_accuracy_Losses_Functions} for the four loss functions, the $k$NN algorithm never led to more than a 2\% average accuracy difference. This performance of the $k$NN algorithm indicates that the DL model learns a suitable input encoding space (i.e., with suitable PPs). 
\begin{table*}[]
\resizebox{\textwidth}{!}{%
\begin{tabular}{lccccc}
\hline
\multicolumn{1}{c}{} &
  \begin{tabular}[c]{@{}c@{}}Acc. with\\      CE Loss\end{tabular} &
  \begin{tabular}[c]{@{}c@{}}Acc. with\\      CE + ICNN \\      Loss\end{tabular} &
  \begin{tabular}[c]{@{}c@{}}Acc. with\\      ProtoPNet Loss\\      ($\epsilon = 0.01$)\end{tabular} &
  \begin{tabular}[c]{@{}c@{}}Acc. with CE + \\      ProtoPNet + \\      ICNN Loss\end{tabular} &
  \begin{tabular}[c]{@{}c@{}}CNN models\\      Acc. With CE\end{tabular} \\ \hline
\multicolumn{1}{r}{KNN on closest feature tensors $z_{w,h}$ to PPs $p_{m,k}$} &
  89.17±5.9\% &
  88.71±4.72\% &
  86.88±8.13\% &
  91.30±0.86 &
  NA \\
Final FC layer &
  89.65±2.11 &
  90.37±0.58 &
  87.67±1.54 &
  90.30±1.84 &
  83.40±5.32 \\ \hline
\end{tabular}%
}
\caption{\textbf{KNN on the learned feature space:} Accuracy obtained for classifying the closest $z_{w,h}$ per image, to the PPs $p_{m,k}$ with a 5-Fold KNN, with K=5. These measurements were performed using the best model obtained for each training loss function. These models had a Resnet50 as the feature extractor and 3 PPs per class.}
\label{table_KNN}
\end{table*}
\begin{figure*}[ht] 
    \centering
    \includegraphics[width=0.9\linewidth]{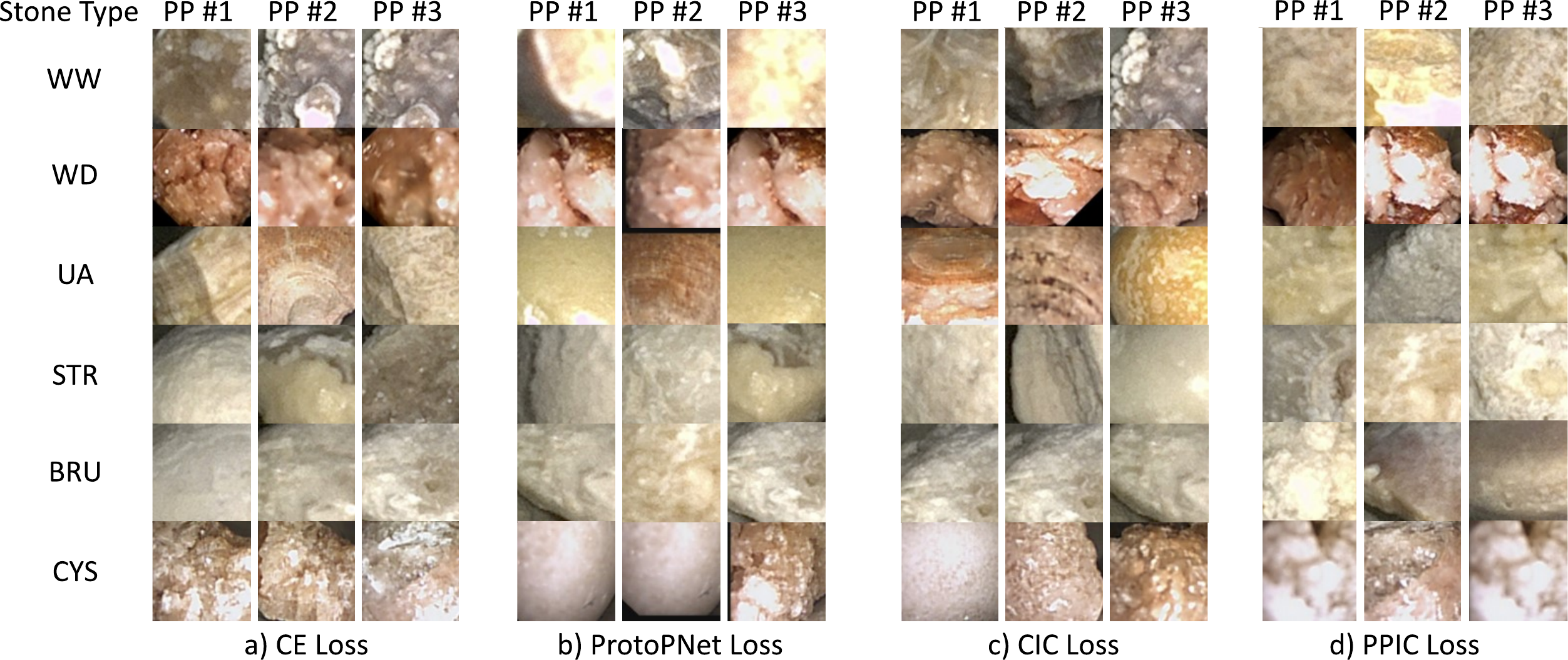}
    \caption{\small{
    PPs learned by models taking ResNet50 as the backbone. Three PPs were learned for each of the six classes for the four losses under study. 
    It is noticeable that the model-training with the {\em CIC loss} favours diversity and leads to representative PPs. 
    }}
    \label{fig_learned_pps}
\end{figure*}

\begin{figure*}[ht] 
    \centering
    \includegraphics[width=1.0\linewidth]{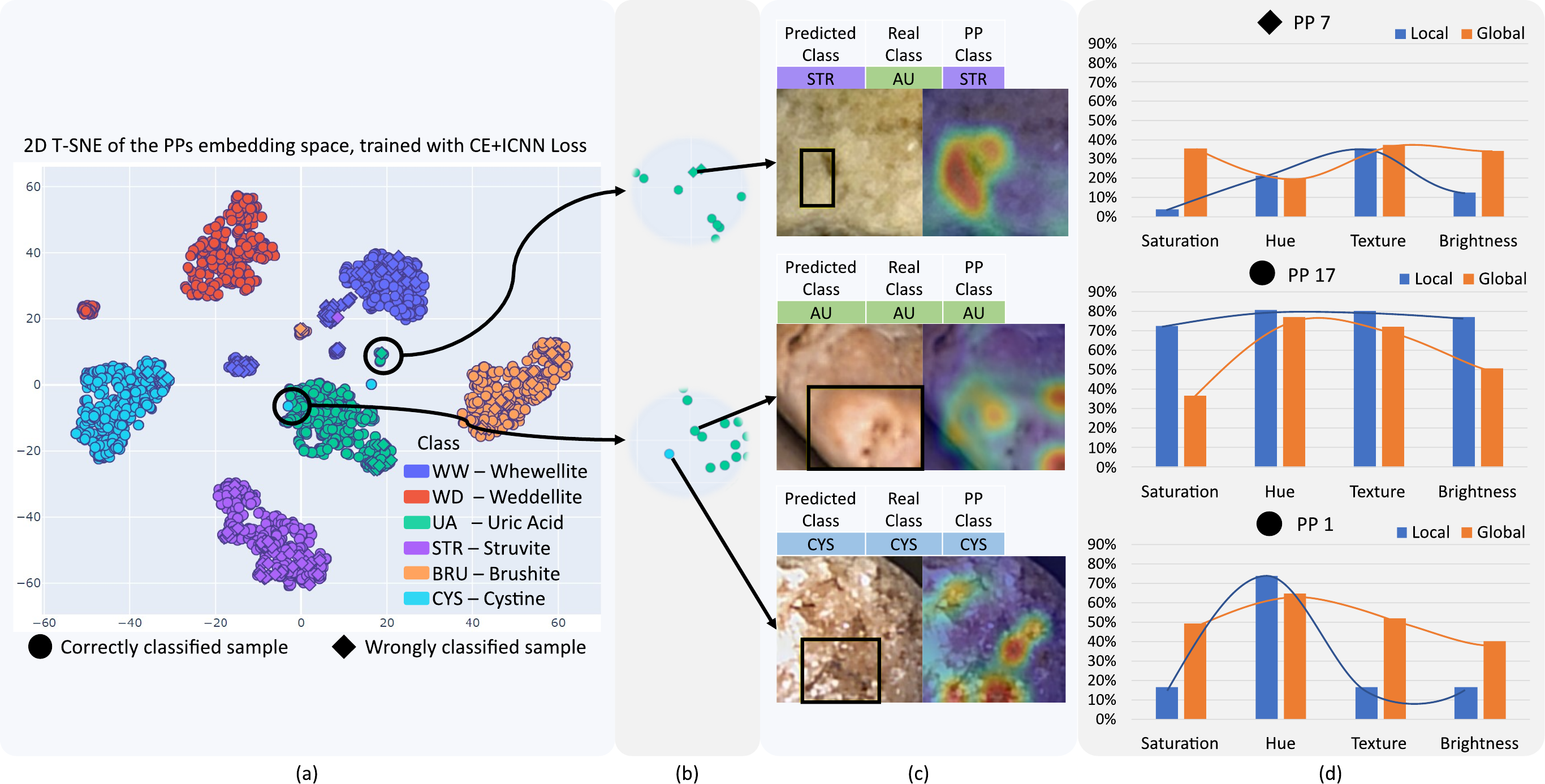} 
    \caption{
    \small{ 
    Illustration of three decisions taken by the proposed model. 
    Subfigure (a) gives the 2D t-SNE space of latent feature tensors $z_{w,h}$ that are the most similar to the classifier's Prototypical Parts (PPs). Those feature activations were extracted from the test kidney stone images.
    This 2D t-SNE space was obtained with a model integrating a ResNet50 backbone, using 3 PPs per class, and trained with the CIC loss function. 
    Subfigure (b) shows two zoom-in areas from the 2D t-SNE map. The green diamond shape sample (at the start of the top arrow) relates to an incorrectly classified input image, whereas the two circles (blue and green circles at the start of the two other arrows) correspond to correctly classified input images. 
    Subfigure (c) displays the predicted class, the real class of the input images, and the class of the closest PP $p_{m,k}$. Additionally, a bounding box delineates the area in the input image with the highest similarity to the closest PP. On the right of the image is the corresponding heat map showing the similarity of the image region and the PP.
    The class label of an input image is given by its most similar PP. 
    Subfigure (d) graph the \textit{descriptors} for the closest PP for each of the three cases in subfigure (c). 
    The example on the top of (d) shows that, when the closest PPs (here PP number 7), don´t have a concordance of its highest \textit{local} and \textit{global} \textit{descriptors} values, it is most likely inducing an incorrect predicted class. 
    Also, the two examples in the middle and bottom of this subfigure present their highest descriptors above the 50\% level. 
    }} 
    \label{fig_2D_TSNE_Classifications}
\end{figure*}

\begin{figure*}[] 
\centering
    \includegraphics[width=0.80\linewidth]{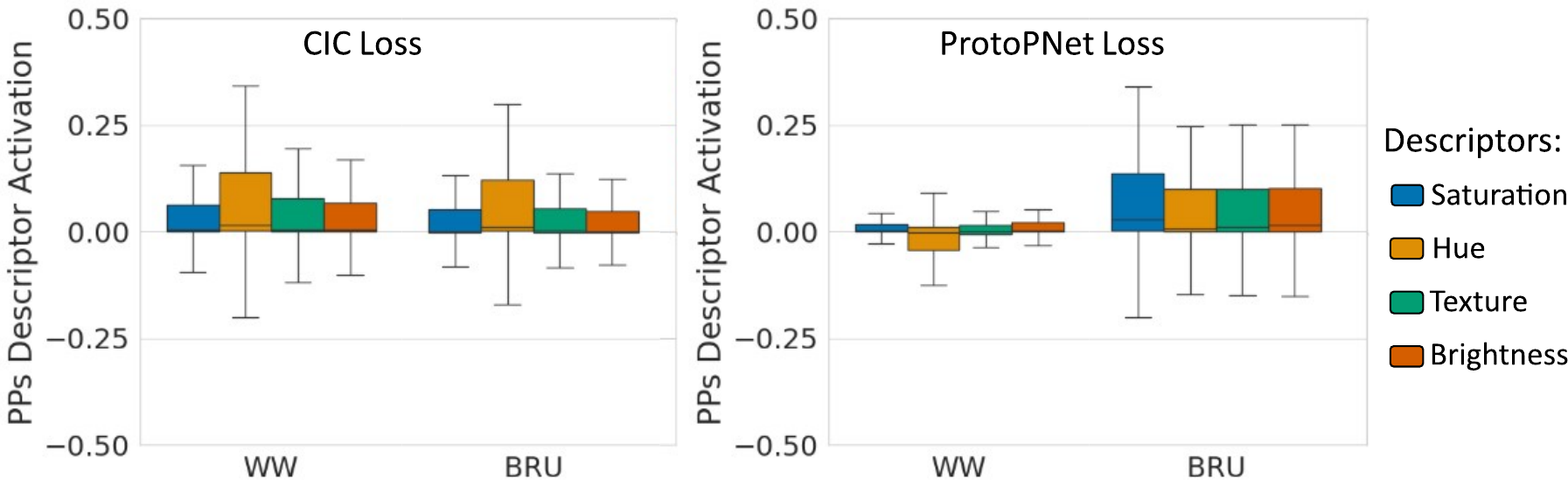} 
    \caption{\small{Activation values given for four PP descriptors and for two classes (i.e., whewellite and brushite types denoted by WW and BRU , respectively). The models leading to these values integrated ResNet50 as the backbone and 3 PPs per class. 
    }} 
\label{fig_Descriptors_per_class_2classes}
\end{figure*}

\vspace{1mm}
{\bf Visual PPs diversity.} Figure \ref{fig_learned_pps} presents the PPs learned using the training set and obtained by the four best DL models using one of the four different loss functions under study. As noticeable in this figure, the loss function choice significantly impacts the visual diversity of the learned PPs, this diversity being the highest for the {\em CIC} loss function $\mathcal{L}_{CIC}$.
The models trained with the \textit{CE loss} (see  Fig.~\ref{fig_learned_pps}.(a)) and the \textit{ProtoPNet loss}  (see  Fig.~\ref{fig_learned_pps}.(b)) produced PPs collapse, as they both learned very similar PPs for each of the classes. 
 This lack of diversity can be appreciated for PP\#$2$ and PP\#$3$ of class WW and for PP\#$1$ and PP\#$2$ of class CYS of the \textit{CE loss} in Fig.~\ref{fig_learned_pps}.(a). Lack of diversity  is also observable  for PP\#$1$ and PP\#$3$  of classes WD and UA, as well as for the PP\#$1$ and PP\#$2$ of class CYS of the \textit{ProtoPNet loss} in Fig.~\ref{fig_learned_pps}.(b).
On the other hand, as seen in Fig.~\ref{fig_learned_pps}.(c), the model trained with the \textit{CIC loss} yielded a more diverse set of PPs. 
For instance, for the CYS-class, the $\mathcal{L}_{CIC}$ loss function led to three PPs, including each texture with a different level of granularity (almost no texture on the left PP, fine-grained textures in the central PP and more roughly grained textures on the right PP). This texture diversity of the CYS-class is not present in the PPs obtained with the other three loss functions.
The qualitative results given in  Fig.~\ref{fig_learned_pps} confirm the quantitative results in Table \ref{table_accuracy_Losses_Functions} (model accuracy) and the FID-scores discussed in Section \ref{Comparative_Quantitative_Analysis}. 

In particular, it can be noticed that the PPs images from models trained with the \textit{CIC loss} achieved a higher visual similarity to the test set than models trained only with the \textit{ProtoPNet loss}, this result being confirmed by the FID scores of 75.88 and 88.12 obtained with the $\mathcal{L}_{CIC}$ and $\mathcal{L}_{ProtoPNet}$ loss functions, respectively. 
These improvements in FID scores and the visual diversity achieved by the model trained with three PPs per class and the \textit{CIC loss} not only indicate a higher diversity of learned PPs, but also suggest a higher similarity to new inference samples. 

Thus, since the \textit{CIC loss} function tends to produce more general PPs, the model producing these PPs has a higher generalization capabilities.   

%
\subsection{Explanations using descriptors  \label{Explanation_visualization}} 
As sketched in Fig.~\ref{fig_2D_TSNE_Classifications}, the aim of this section is to illustrate the behavior of the proposed model using a case-based method and descriptors. The three classification results analyzed in detail below (two correct classifications and a misclassification, see Fig.~\ref{fig_2D_TSNE_Classifications}) were selected in a 2D t-SNE visualization generated by the best-performing model incorporating a \textit{ResNet50} backbone, based on three PPs per class and trained with the $\mathcal{L}_{CIC}$ loss function.

The model behavior illustration starts with the analysis of an incorrect model decision for which a tensor patch $z_{w,h}$ of the uric acid (UA) kidney stone type led to the identification of a struvite (STR) renal calculi type.
This misclassification is evidenced by the dissimilar local descriptor (blue bars) activation values against the global descriptors (orange bars) activation values of same PPs, particularly  (see the example on the PP\#7 graph at the top of Fig.~\ref{fig_2D_TSNE_Classifications}.(d)). 
The analysis reveals that no descriptor (\textit{local} or \textit{global}) reaches over 40\% activation, indicating a lack of visual similarity between PP\#7 and the input image.
Conversely, the middle and bottom graphs in Fig.~\ref{fig_2D_TSNE_Classifications} represent correctly classified images. For instance, a Cystine (CYS) kidney stone image, depicted with a blue circle, is correctly identified but located near a different class cluster. The relevance of the distinct visual characteristics, as demonstrated in the right-sidebar graphs, validates its classification. 
This is, this case presents high \textit{local} and \textit{global} descriptor activations, which correspond to the highest and second highest values. One can observe that, for correct classifications, the \textit{local} descriptors of a PP $p_{m,k}$, follow the same trend in values as its \textit{global} descriptors for a same PP $p_{m,k}$.

Further indications of the model behavior can be deduced from Fig.~\ref{fig_Descriptors_per_class_2classes}. This figure shows the discrimination power of visual features through four descriptors of the PPs of two kidney stone types, namely whewellite (WW) and brushite (BRU). 
For an accurate classification, it is important that the sum of the local scores of the PPs descriptors of the complete training set of a class reveals a hierarchy (or a difference) in the descriptor relevance. Indeed, the classification is usually less accurate when all descriptors highlight the similar importance of visual features in the decision process. 
As noticeable in the left part of  Fig.~\ref{fig_Descriptors_per_class_2classes}, the model that uses the $\mathcal{L}_{CIC}$ loss function tends to produce a clear hierarchy in terms of descriptor importance since for both the WW and the BRU types the hue, texture, brightness, and saturation features led to descriptor activation values that can be ranked according to a decreasing importance in terms of discrimination ability. Such a differentiation in visual feature importance indicates a high model accuracy. On the contrary, for the \textit{ProtoPNet loss} (see the right part of  Fig.~\ref{fig_Descriptors_per_class_2classes}), three of the four visual features exhibit similar PPs descriptor activation values. Only the hue and saturation have slightly different importance for the WW and BRU types, respectively. Such uniformly distributed importance of the PPs descriptor activations indicates a low model performance.
In practice, for the kidney stone type recognition, the \textit{CIC loss} has an accuracy from 2\% up to 4\% higher than that obtained with the {\em ProtoPNet loss}.

\section{Discussion \label{Discussion}} 
This contribution aims to improve the explainability of a reference DL model.
The proposed approach modified the ProtoPNet training to enhance its interpretability. 
Various training parameters, including the backbone type, the number of PPs per class, and different loss functions, were explored to identify optimal network configurations.
Notably, the proposed approach maintains the characteristic of forgoing part annotation for training, relying solely on class labels. 
This avoids requiring additional efforts from specialists in the generation of the dataset since the current classification used for training is already part of the current process to attend the patients.
The results showed that modifying and fine-tuning the model training improves it by providing more detailed and faithful explanations while maintaining the same level of accuracy. 

Additionally, Figs.~\ref{fig_TSNE_comparison} and \ref{fig_learned_pps} demonstrate that incorporating a DML approach (as the ICNN-score-based loss) during the training, enhances the accuracy due to an effective clusterization (see Fig.~\ref{fig_TSNE_comparison}(c)) and lead to a higher diversity in terms  of texture granularity of the learned PPs (see Fig.~\ref{fig_learned_pps}(c)). 
The accuracy results and FID scores support the idea that refining the loss function characteristics, which guide the training, leads to better extraction of latent feature tensors $z_{w,h}$ and of the PPs $p_{m,k}$ in the high dimensional latent space. This, in turn, improves intra-class diversity. 

It was found that including variance estimation, with the $\Omega$ function, which is a part of the $\mathcal{L}_{\mathrm{ICNN}}$ loss, produced clusters with better continuity. This means that the learned clusters had more constant distances between data points, and most of the same class points were continuous with each other within each cluster. 
This improvement was observed when comparing the ICNN loss (see Fig.~\ref{fig_2D_TSNE_Classifications}(c)) to the ProtoPNet loss (see Fig.~\ref{fig_2D_TSNE_Classifications}(b)) and the conventional Cross-Entropy loss (Fig.~\ref{fig_2D_TSNE_Classifications}(a)). The ICNN loss not only maintains diversity, but also enhances the structural integrity of clusters in the latent space, leading to more representative and coherent prototypes.
he framework proposed in Fig.~\ref{fig_general_architecture} allows for the encoding of the main visual features from an input image into latent feature tensors $z_{h,w}$. The similarity level of the latter to the prototypical cases $p_{m,k}$ of each class $k$ is automatically measured and visually located on input image $x_n$ with similarity heatmaps $\mathbf{H}_{m,k}$. An example of one of those heatmaps $\mathbf{H}_{m,k}$ is shown in Fig.~\ref{fig_PP_similarity_visualization}. 
Additionally, the global photometric perturbation can also be extracted, as shown in Fig.~ \ref{fig_Descriptors_per_class_2classes}, giving an explanation per class of the main visual features the model learned to recognize each kidney stone type. 
This case-based reasoning approach generates explanations and additional activation details, allowing experts to use these models as an assistance tool for the MCA, since the heatmaps indicate ``where'' the model is looking at to classify an image. The image example associated with each PP $p_{m,k}$ (as seen in Fig.\ref{fig_learned_pps}) gives a visual illustration of what other instances the model found most similar so that the medical specialist can compare if those cases are adequate according to their expertise. 
Additionally, the local descriptors highlight which visual features and the degree of intensity with which the model identified similarities between the input image and the prototypical parts (PPs). By comparing these descriptors at both the local and global levels, it becomes possible to assess whether the similarity between a PP and an input image aligns with the patterns learned during training, as explained in the caption of Fig. \ref{fig_2D_TSNE_Classifications}.
%
More importantly, this contribution presents the evidence of how to use the generated explanations and descriptors for correct and incorrect cases. For instance, Fig.~\ref{fig_2D_TSNE_Classifications} gives an example of how to analyze the overall behavior of class descriptors
to identify the reasons behind misclassification.
PPs models inherit most of the performance characteristics of CNNs. This suggests that methods to enhance CNN performance should also improve PPs models, thus expanding potential improvements. 

While the {\em CIC loss} significantly improves the clustering of class samples, as evidenced in the t-SNE visualization in Fig.~\ref{fig_2D_TSNE_Classifications}(c), some outliers remain. This observation suggests that even if  the CIC loss $\mathcal{L}_{CIC}$ enhances intra-class compactness and inter-class separation, further refinement may be necessary to avoid all outliers. Outliers in t-SNE visualizations are not uncommon due to the dimensionality reduction process, which can sometimes distort local relationships. However, the overall improvement in clustering quality underscores the effectiveness of the CIC loss $\mathcal{L}_{CIC}$ in achieving better intra-class compactness and inter-class separability compared to other loss functions.
As observed for all class clusters, except for the whewellite (WW) type in Fig.~\ref{fig_2D_TSNE_Classifications}(c), most of the outliers were correctly classified. This leads to a more general and human-interpretable model behavior.

Table \ref{table_accuracy_Losses_Functions} shows that assigning a higher number of PPs per class does not significantly improve the classification performance of PPs-based models. On the contrary, a small number of PPs per class facilitates the interpretation of specialists observing the areas of interest used and detected by this model type.  
Also, regarding the diversity of the learned PPs (see Fig.~\ref{fig_learned_pps}), it is important to have an adequate diversity of visual feature values in the learned PPs. This helps to generate explanations of image classifications that help urologists or biologists to recognize kidney stones based on several features.
In other words, the model identifies and focuses on a few class features that biologists or urologists can interpret.

 The proposed model also addresses the oversimplifications common in current XAI visualization methods, particularly in the context of kidney stone classification. It utilizes a case-based reasoning process to extract semantic features from input images using a CNN.
Furthermore,  descriptors are used to quantify the sensitivity of PPs to various visual perturbations. This allows for an in-depth analysis of the significance of visual features for each PP. The granularity in explanation aids specialists in understanding the underlying reasoning behind the model's output.

Finally, the conducted $k$NN tests revealed interesting observations. 
These results showed that the latent feature tensors $z_{w,h}$ extracted by the model trained with \textit{PPIC loss}, got a 91\% accuracy with the $k$NN algorithm, performing better by 1\% compared to the original model architecture which uses an FC-connected classifier layer.
This result shows that an optimization of the clusterization characteristics in terms of  inter- and intra-class distances between the latent feature tensors $z_{w,h}$ and  learned PPs $p_{m,k}$ improves the classification performance.
 Further tests will be conducted to confirm this observation. 
 By limiting the number of prototypical parts, the model facilitates user comprehension and exhibits competitive performance compared to its non-interpretable counterparts. This approach closely aligns with the methods used by medical specialists, thereby enhancing the relevance and utility of the explanations provided.
It was shown that by adapting and fine-tuning CNN models into ProtoPNets and adjusting training with a deep metric learning approach, specifically with the ICNN score, it is possible to transform black-box CNN models into self-explainable ones and increase their alignment with medical experts, providing detailed and faithful explanations.
\section{Conclusion and further work}\label{Conclusion}

This contribution presents a framework that provides a detailed, interpretable, and reliable explanation of the visual features behind misclassifications and correct classifications, as shown in Figure \ref{fig_2D_TSNE_Classifications},  which are relayed upon by specialists to classify kidney stones.
The proposed model provides valuable insights to assist specialists in their diagnostic processes and foster trust in AI systems. 
This trust is essential for the effective integration of AI in healthcare, enabling specialists to verify AI outputs for accuracy and plausibility and, if necessary, to override them with their expert judgment. Our approach paves the way for an increment in the adoption of DL models by medical experts as tools for their medical diagnostics, where AI and human expertise could began to collaborate to achieve better patient outcomes.

Further work will explore the tuning and modification of the ICNN loss function. The traditional approach of penalizing close PPs from different classes has been shown to be effective in learning useful PPs for classification. However, to prevent PPs of the same class from collapsing, it is necessary to encourage proximity while maintaining a minimum margin distance between them and a minimum variance sustained by each class cluster.  
Also, future research could refine the model's reasoning and reporting mechanisms to align more closely with the structured lexicon used by urologists.
This could be achieved by incorporating natural language descriptions for the PPs identified and their \textit{descriptors} to make explanations more accessible to specialists and a broader range of users.

\section*{Acknowledgments}
The authors wish to acknowledge the Mexican Council for Science and Technology (CONAHCYT) for their support in terms of postgraduate scholarships in this project. 

This work has been supported by Azure Sponsorship credits granted by Microsoft's AI for Good Research Lab through the AI for Health program. We thank the Data Science Hub at Tecnologico de Monterrey for making this possible.

The project was also supported by the French-Mexican ANUIES CONAHCYT Ecos Nord grant 322537.
The authors would like to thank the financial support from Tecnologico de Monterrey through the “Challenge-Based Research Funding Program 2022”. Project ID 	\# E120 - EIC-GI06 - B-T3 - D.

\section*{Compliance with ethical approval}
The images were captured in medical procedures following the ethical principles outlined in the Helsinki Declaration of 1975, as revised in 2000, with the consent of the patients.

\bibliographystyle{cas-model2-names}

\bibliography{cas-refs}

\appendix
\section{Appendix A: ICNN Score Design}\label{Appendix_A}
\begin{figure}[t] 
\centering
    \includegraphics[width=1.0\columnwidth]{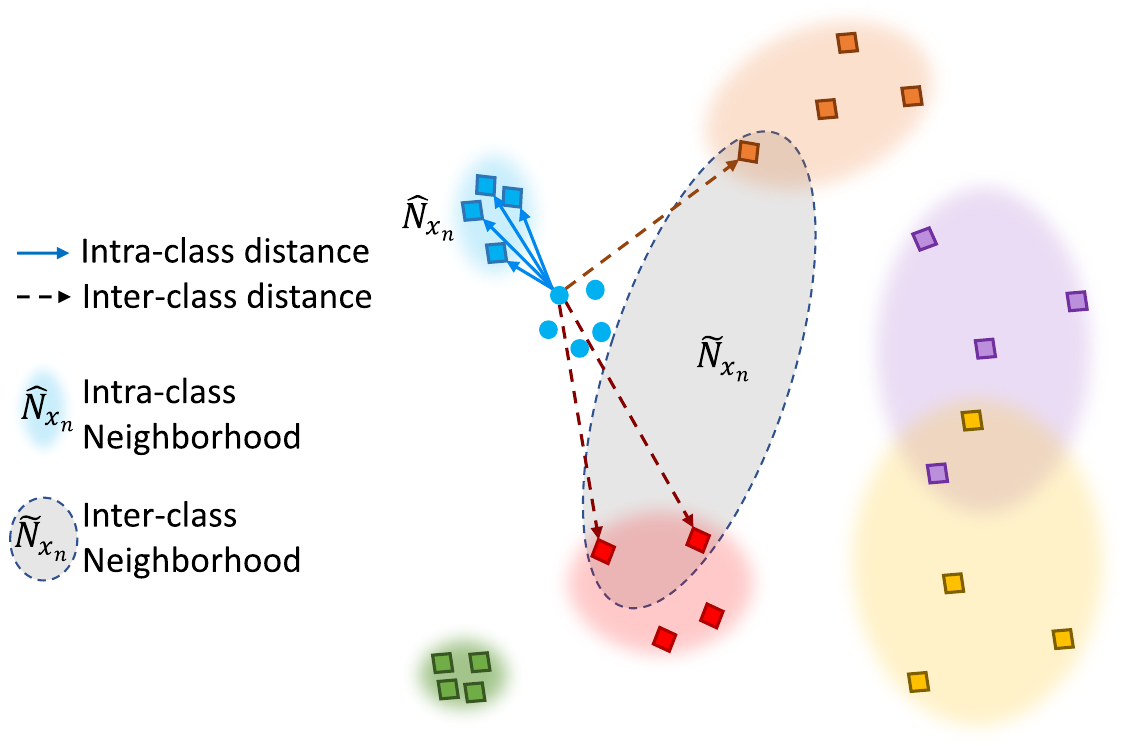} 
    \caption{\small{
    Intra- and inter-class representation in the latent feature space.
    The dark blue squares in the light blue elliptical area represent the PPs  $p_{m,k}$ of the Intra-class neighborhood $\widehat{N}_{x_n}$, with $|\widehat{N}_{x_n}=4|$. These PPs are the nearest to the input image feature tensors $z_{w,h}^k$ corresponding to the blue circles.   
    The elliptical grey area (delineated by a dashed line) defines inter-class neighborhood $\widetilde{N}_{x_n}$ including $|\widetilde{N}_{x_n}=3|$ PPs of two others classes located in the orange and red ellipses.
    It is noticeable that the blue and green elliptical areas present a lower PPs diversity than that in the purple and yellow areas. 
    }} 
\label{fig_ICNN_PPs_embeding}
\end{figure}

%
Functions $\Lambda$, $\Omega$ and $\Gamma$ used in Eq.~\eqref{eq_ICNN} which must be maximized are determined in the latent feature space using the L2 distances between tensors $z_{w,h}^k$ and PPs $p_{m,k}$, and are based on the concept of intra- and inter-class neighborhoods sketched in Fig. \ref{fig_ICNN_PPs_embeding}:   
\begin{itemize}
\item An intra-class neighborhood $\widehat{N}_{x_n}$ is defined by the set of the $|\widehat{N}_{x_n}|$ prototypical parts $\widehat{p}_{m,k}$ being the nearest to a convolutional feature tensor $z_{w,h}^k$ extracted from image $x_n$ and belonging to the same class $k$ as that of its $|\widehat{N}_{x_n}|$ closest PPs. 
\item In contrast, an inter-class neighborhood $\widetilde{N}_{x_n}$ consists of the $|\widetilde{N}_{x_n}|$ PPs $\widetilde{p}_{m,k}$ ($k \neq i$) which are the nearest to a convolutional feature tensors $z_{w,h}^{k}$ extracted from image $x_n$ and belonging to class $k = i$. 
\end{itemize}
%
The distances $h(z_{w,h}^k, p_{m,k})$ used to measure the proxi\-mi\-ty between tensors $z_{w,h}^{k}$ and prototypes $p_{m,k}$ 
are norma\-lized distances $d(z_{w,h} ^k, p_{m,k})$. In Eq.~\eqref{eq:dnormalization}, the normalization is performed with $\theta(x_n)$ and $\alpha(x_n)$, which respectively stand for the smallest and largest possible distance between a prototype $p_{m,k}$ and a latent feature tensor $z_{w,h}^{k}$. These min-max values are determined for a given feature tensor $z_{w,h}^k$ and the prototypes $p_{m,k}$ located in neighborhood $\widehat{N}_{x_n} \cup \widetilde{N}_{x_n}$.  
\begin{equation} \label{eq:dnormalization} 
 h\left(z_{w,h}^{k}, p_{m,k} \right) = \frac{d\left(z_{w,h}^{k}, p_{m,k} \right) - \theta\left( x_n \right)}{\alpha\left( x_n \right) - \theta\left(x_n\right)} \quad \textrm{with} \qquad 
 \end{equation} 
 \vspace*{-3mm}
 \begin{equation} 
  \left\{
      \begin{aligned}
      \alpha \left( x_n \right)&=&\max \left\{ d\left( z_{w,h}^{k}, p_{m,k} \right) \vert p_{m,k} \in \widehat{N}_{x_n} \cup \widetilde{N}_{x_n} \right\} \\
      \theta \left( x_n \right)&=&\min \left\{ d\left(  z_{w,h}^{k}, p_{m,k} \right) \vert p_{m,k} \in \widehat{N}_{x_n} \cup \widetilde{N}_{x_n} \right\}
      \end{aligned} \nonumber
  \right.
\end{equation} 
This min-max normalization constrains $h(z_{w,h} ^k, p_{m,k})$ within the distance range $[0,1]$. 
The $\lambda_{\text{inter}}$-value defined in Eq.~\eqref{TotalNormDistance} corresponds to the sum of all distances between a given latent feature tensor $z_{w,h}^{k}$ of class $k=i$ and the $|\widetilde{N}_{x_n}|$ closest prototypes $\widetilde{p}_{m,k}$ belonging to classes $k \ne i$ included in set $\widetilde{N}_{x_n}$.
In a similar way, the $\lambda_{\text{intra}}$-value given in Eq.~\eqref{TotalNormDistance} is the sum of all distances between a given latent feature tensor $z_{w,h}^{k}$ of class $k=i$ and the $|\widehat{N}_{x_n}|$ closest prototypes $\widehat{p}_{m,k}$ of class $k=i$. These intra- and inter-class values are used to build the three $\Lambda$, $\Omega$ and $\Gamma$ functions. 
\begin{equation} 
  \left\{
\begin{aligned}
    \lambda_{\text{inter}} \left(z_{w,h}^{k},\widetilde{p}_{m,k} \right) & = \sum_{m=1}^{m=|\widetilde{N}_{x_n}|} h \left(  z_{w,h}^{k}, \widetilde{p}_{m,k} \right) \label{TotalNormDistance} \\
    \lambda_{\text{intra}} \left( z_{w,h}^{k}, \widehat{p}_{m,k}\right) & = \sum_{m=1}^{m=|\widehat{N}_{x_n}|} \left[ 1-h\left(z_{w,h}^{k}, \widehat{p}_{m,k} \right) \right]   
\end{aligned}
  \right.
\end{equation}
It is noticeable that the range of possible values of $\lambda_{\text{inter}}$ is $[0, |\widetilde{N}_{x_n}|]$, where $\lambda_{\text{inter}}(z_{w,h}^{k}, \widetilde{p}_{m,k}) = |\widetilde{N}_{x_n}|$ corresponds to the best separability of latent feature tensor $z_{w,h}^{k}$ of class $k=i$ with prototypes $\widetilde{p}_{m,k}$ of classes $k\neq i$. In the same way, the values of $\lambda_{\text{intra}}$ are in $[0, |\widehat{N}_{x_n}|]$ and a value of $\lambda_{\text{intra}}(z_{w,h}^{k}, \widehat{p}_{m,k})$ tending towards $|\widehat{N}_{x_n}|$ indicates with a very high probability that tensor $z_{w,h}^{k}$ belongs to class $k=i$.      
%

The maximisation of function $\Lambda$ given in Eq.~\eqref{eq:Lambda} implies that the distances between $z_{w,h}^{k}$ and the PPs in set $\widehat{N}_{x_n}$ become all weak, while the distances between $z_{w,h}^{k}$
and the PPs in set $\widetilde{N}_{x_n}$ tend to be simultaneously large.
\vspace{-0.1cm}
\begin{align} \label{eq:Lambda}
 \hspace*{-8mm}   \Lambda \left(z_{w,h}^{k}, p_{m,k}\right)  & = \\ &\frac{1}{2} \left(\frac{\lambda_{\text{inter}} \left( z_{w,h}^{k}, \widetilde{p}_{m,k} \right)}{\left|\widetilde{N}_{x_n}\right|} + \frac{\lambda_{\text{intra}} \left( z_{w,h}^{k}, \widehat{p}_{m,k} \right)}{\left| \widehat{N}_{x_n}  \right|} \right) \nonumber
\end{align}
The value of function $\Lambda$ depends on the distances between a feature tensor $z_{w,h}^k$ and all prototypes $p_{m,k}$ in $\widehat{N}_{x_n} \cup \widetilde{N}_{x_n}$. These values range in [0, 1]. $\Lambda$ values approaching the maximal value of 1 favour two effects: one the one hand, the PPs learned for each class $k$ tend to form compact clusters, and on the other hand, the distances between these clusters increase to allow for class separability in the latent feature space. Thus, the maximization of function $\Lambda$ used in Eq.~\eqref{eq_ICNN} facilitates the classification, i.e., the kidney stone type recognition.
 %
%
%

However, increasing the class compactness also tend to diminish the class diversity carried by the prototypical parts, which involves the risk of PPs collapse. A solution to minimize the risk of PPs collapse is to select in class $k$ the $|\widehat{N}_{x_n}|$ PPs leading to the greatest variability of the distances $h(z_{w,h}^{k}, \widehat{p}_{m,k})$ in set $\widehat{N}_{x_n}$. Variance $Var_{\textrm{intra}}$ given in Eq.~\eqref{eq:intraClassVariabilty} is a measure of the intra-class PPs diversity.
\begin{align} \label{eq:intraClassVariabilty} 
 \hspace*{-9mm}Var_{\textrm{intra}}\left(z_{w,h}^{k}, \widehat{p}_{m,k} \right) & = \\ & 
 \frac{1}{\widehat{N}_{x_n}} \sum_{m=1}^{m=|\widehat{N}_{x_n}|}\left(\overline{h}-h\left(z_{w,h}^{k}, \widehat{p}_{m,k} \right)\right)^2
 \nonumber
 \end{align} 
 \vspace*{-3mm}
 \begin{equation} 
\qquad  \textrm{with} \quad \overline{h} = \frac{1}{\widehat{N}_{x_n}} \sum_{m=1}^{m=|\widehat{N}_{x_n}|} h\left(z_{w,h}^{k}, \widehat{p}_{m,k} \right)  \nonumber
\end{equation} 
In the latent feature space, the separability of the cluster of class $k=i$ should also be simultaneously ensured with all other clusters of classes $k\neq i$. This separability can be obtained  by maximizing the diversity in terms of prototypes in set $\widetilde{N}_{x_n}$ (i.e., PPs from different classes $k\neq i$ must belong to set $\widetilde{N}_{x_n}$). This inter-class diversity can again be formulated as a variance. In Eq.~\eqref{eq:interClassVariabilty} giving this variance, ratio $\lambda_{\text{inter}}(z_{w,h}^{k}, \widetilde{p}_{m,k})/|\widetilde{N}_{x_n}|$ corresponds to the mean of the distances $h(z_{w,h}^{k}, \widetilde{p}_{m,k})$ determined with all $ \widetilde{p}_{m,k} \in \widetilde{N}_{x_n}$.
\begin{align} \label{eq:interClassVariabilty} 
 \hspace*{-9mm}Var_{\textrm{inter}}\left(z_{w,h}^{k}, \widetilde{p}_{m,k} \right) & = \\ &  \hspace*{-23mm} 
 \frac{1}{\widetilde{N}_{x_n}} \sum_{m=1}^{m=|\widetilde{N}_{x_n}|}\left(\frac{\lambda_{\text{inter}}\left(z_{w,h}^{k}, \widetilde{p}_{m,k}\right)}{|\widetilde{N}_{x_n}|}-h\left(z_{w,h}^{k}, \widetilde{p}_{m,k} \right)\right)^2
 \nonumber
 \end{align} 
 It is noticeable that $\lambda_{\textrm{inter}}$ maximizes the sum of the distances between class $k=i$ and classes $k\neq i$ (some distances can remain small), while the maximization of $Var_{\textrm{inter}}$ contributes to a simultaneous increase of these distances. 

 The intra- and inter-class diversity are maximized using function $\Omega$ defined in Eq.~\eqref{diversity maximization}.
\vspace{-0.1cm}
\begin{align} \label{diversity maximization}
 \hspace*{-9mm} \Omega \left(z_{w,h}^k,p_{m,k}\right) = \\ &
   \hspace*{-12mm}  Var_{\textrm{intra}}\left(z_{w,h}^{k}, \widehat{p}_{m,k} \right) + Var_{\textrm{inter}} \left(z_{w,h}^{k}, \widetilde{p}_{m,k}  \right) \nonumber
\end{align}
In Eq.~\eqref{eq_ICNN}, function $\Omega$ modulates (or penalizes) function $\Lambda$ so that the search of cluster compactness and large class distances are done while limiting PPs collapse risks and ``unbalanced'' class distances.  




Finally, the last component of the ICNN Score takes into account the fact that the PPs included in sets $|\widehat{N}_{x_n}|$ and $|\widetilde{N}_{x_n}|$ are the closest to latent feature tensor $z^k_{w,h}$. When a such a tensor is assigned to class $k=i$, then most of his closest PPs neighbors should belong in class $k=i$ while the closest PPs neighbors belonging to classes $k\neq i$ should be less numerous (i.e., for all $z^k_{w,h}$ on should have $|\widehat{N}_{x_n}| > |\widetilde{N}_{x_n}|$).
This idea is mathematically formulated in Eq.~\eqref{gamma_func}. It is worth noticing that the sum $|\widehat{N}_{x_n}| + |\widetilde{N}_{x_n}|$ is constant.
\vspace{-0.1cm}
\begin{equation}\label{gamma_func}
    \Gamma \left(z_{w,h},p_{m,k}\right) = \frac{\left| \widehat{N}_{x_n}  \right|}{\left| \widehat{N}_{x_n}  \right|+\left| \widetilde{N}_{x_n}  \right|}
\end{equation}

Maximizing function $\Gamma \left(z_{w,h},p_{m,k}\right)$ by choosing appropriate PPs for all classes helps to  reinforce the accuracy of the classification. In Eq.~\eqref{eq_ICNN}, function $\Gamma(z_{w,h},p_{m,k})$ modulates product $\Lambda(z_{w,h},p_{m,k})\times \Omega(z_{w,h},p_{m,k})$ and high values of $\Gamma$ favour solutions for which a latent feature tensor is close to a large number of PPs of its class.  

%


\end{document}